\documentclass[11pt]{article}

\usepackage[final]{acl}

\usepackage{times}
\usepackage{latexsym}

\usepackage[T1]{fontenc}
\usepackage[utf8]{inputenc}

\usepackage{microtype}

\usepackage{inconsolata}

\usepackage{graphicx}

\usepackage{amsmath}
\usepackage{amsfonts}

\usepackage[table]{xcolor}
\usepackage{colortbl}   
\usepackage{tabularx}     
\usepackage{xfp}
\usepackage{array}

\usepackage{etoolbox}

\usepackage{booktabs}
\usepackage{multirow}
\usepackage{hhline}
\usepackage{comment}

\newcolumntype{Z}{>{\centering\arraybackslash}X}

\definecolor{pale_green}{RGB}{50, 150, 200}

\newcommand{\colorcellinner}[2]{%
  \cellcolor{pale_green!#1}#2%
}

\newcommand{\colorcell}[1]{%
  \colorcellinner{%
    \fpeval{round(min(100, ((#1)/1.7)*100), 0)}
  }{#1}%
}

\newcommand{\colorcellentropy}[1]{%
  \colorcellinner{%
    \fpeval{round(min(100, ((#1)/0.6)*100), 0)} 
  }{#1}%
}

\title{Beyond Early-Token Bias: Model-Specific and Language-Specific Position Effects in Multilingual LLMs}

\author{
Menschikov Mikhail\textsuperscript{*} \\
ITMO \\
Applied AI \\
\And
Alexander Kharitonov\textsuperscript{*} \\
SberAI\\
\And
Maiia Kotyga \\
Applied AI \\
\And
Vadim Porvatov \\
Sber \\
\AND
Anna Zhukovskaya \\
Lomonosov MSU \\
\And
David Kagramanyan \\
HSE University \\
\And
Egor Shvetsov\textsuperscript{\textdagger} \\
Applied AI \\
\And
Evgeny Burnaev \\
Applied AI \\
AIRI \\
}

\begin{document}
\maketitle
\begin{abstract}
Large Language Models (LLMs) exhibit position bias systematically underweighting information based on its location in the context but how this bias varies across languages and models remains unclear. We conduct a multilingual study across five typologically diverse languages (English, Russian, German, Hindi, Vietnamese) and five model architectures, analyzing how position bias interacts with prompting strategies and affects output entropy. Our key findings are:   \textbf{(1)} Position bias is primarily model-driven but shows language-specific nuances. Notably, Qwen2.5-7B-Instruct, DeepSeek 7B Chat and Mistral 7B consistently favor late positions challenging the common assumption of universal early-token preference.  \textbf{(2)} Explicitly instructing the model, in the presence of irrelevant distractors,  that “\textit{the most relevant context to the query is marked as \textbf{1}}” unexpectedly reduces accuracy across all languages, questioning standard prompt-engineering practices.  \textbf{(3)} Accuracy consistently drops most  when relevant information appears in the middle of the context, yet this is not reflected in a corresponding increase in output entropy, suggesting the model remains confident even when it fails to use mid-context cues.
\end{abstract}

\section{Introduction}

Many recent applications based on large language models (LLMs) require support for long-context processing\footnote{Such applications may include Retrieval-Augmented Generation, autonomous agents, scientific research, customer support, among others.}. However, developing new training strategies to accommodate longer contexts is insufficient, as new challenges continue to arise. One notable issue is \textbf{position bias} -- the systematic neglect of information located at specific positions, typically in the middle of the context~\citep{baker2024lost}

While position bias has been well-documented in English-centric studies~\citep{zhang2024found,baker2024lost}, its manifestation in multilingual contexts and across-various architectures remains underexplored. Furthermore, existing bias mitigation strategies~\citep{peysakhovich2023attention,zhao2021calibrate,zhang2024can,guo2024serial} have predominantly been evaluated on English-language datasets. As discussed in Section~\ref{sec:cultural_interplay}, lexical and morphological variations across languages necessitate careful consideration in multilingual modeling. As a step toward adapting multilingual large language models for real-world long-context applications, \textit{\textbf{this work seeks to address the following research questions:}}

\noindent\textbf{Q1:} Is position bias primarily a model-driven phenomenon, or do language-specific patterns emerge due to lexical, morphological, and syntactic differences~\citep{ghosh2024morphology}? 

% \noindent\textbf{Q2:} Does position bias cause models to favor English-like SVO (Subject, Verb, Object) word order when generating text in languages with other dominant word orders?
%By answering these questions, we can prioritize directions for position bias mitigation methods. 

\noindent\textbf{Q2:} Do prompt-based strategies, e.g., explicit position guidance~\citep{zhang2024can}, effectively mitigate bias across languages? Can we focus model attention via prompting, and would it improve model performance? 
%This question has a particular interest in Chain of Thought (CoT) strategies, where an explicit positional guidance may be used and for Retrival Augmented Generation (RAG), where context relevancy scoring or reordering is applied, we discuss it further in Section~\ref{sec:rag_reorder}. 

\noindent\textbf{Q3:} Finally, we are interested in a deeper understanding of how position bias affects model generation, and thus we perform formal and empirical analysis of how position bias affects entropy of the output distribution.  

\textbf{Contributions.} We present a multilingual analysis of position bias across five typologically diverse languages (English, German, Russian, Hindi, and Vietnamese) and five model architectures. Our analysis is based on 2,000  sampled examples per language, where each model was evaluated under nine distinct experimental conditions (3 context positions × 3 scoring strategies), yielding 18,000 model generations and evaluations per language. Across all five languages and five models, this totals 450,000 evaluated question–answer pairs.

\textbf{Our key contributions are:}
\begin{itemize}
\item We show that position bias is predominantly model-driven, yet exhibits significant language-specific variations. For instance, models like Qwen2.5-7B-Instruct~\citep{qwen2025qwen25technicalreport}, DeepSeek-7B-Chat~\citep{deepseek-llm} and  Mistral 7B display a strong late-position bias, contradicting prior claims of an inherent early-token preference in LLMs~\citep{wu2025emergence,barbero2025llms}. In contrast, Llama3.1-8B-Instruct~\citep{touvron2023llamaopenefficientfoundation} prioritizes early positions. We speculate these differences stem from variations in training data and model architecture.
\item We find that explicitly instructing models about correct context placement (e.g., “the correct context has label 1”) \textbf{consistently degrades accuracy} across all languages. This result holds when the incorrect context is sampled randomly, a setup that differs from \citet{zhang2024can}, who report that instructions can mitigate position bias. The key distinction is that \citet{zhang2024can} use semantically relevant distractors, whereas our distractors are random.
\item Through an empirical analysis of output entropy, we reveal a counterintuitive dynamic: while model accuracy is lowest when relevant context is in the middle, this performance drop is \textbf{not accompanied by a corresponding peak in entropy}. This indicates a disconnect between positional disadvantage and the model's expressed uncertainty.
% \item While we do not find any connection between position bias and word order, we observe some model-driven patterns.
\end{itemize}
\noindent These findings yield the following \textbf{practical implications}:
\begin{enumerate}
\item Chain-of-Thought and other reasoning strategies that rely on explicit positional guidance demand careful reconsideration, as our results show they can cause consistent performance degradation.
\item Retrieval-augmented generation (RAG) systems should account for specific model and language characteristics. Strategies that reorder context based on the assumption that models prioritize recent or initial tokens (see Section~\ref{sec:rag_reorder}) may be ineffective or detrimental for models with a late-position bias.
\item The observed entropy dynamics complicate uncertainty-based bias mitigation techniques~\citep{duan2023shifting} and are crucial for developing effective uncertainty quantification strategies in RAG pipelines.
\end{enumerate}

The code for our experiments is available in the \href{https://anonymous.4open.science/r/UncertaintyPosition}{GitHub repository}.

\section{Related Work}
\textbf{What causes position bias?} Prior work identifies multiple contributing factors. \citep{zhang2024found} attribute position bias to U-shaped \textit{attention patterns} in transformers, which prioritize extremal positions and degrade performance for mid-context evidence. Theoretical and empirical studies further demonstrate that transformers \textit{attention is inherently biased toward earlier} tokens~\citep{wu2025emergence, barbero2025llms}. \citep{wu2025emergence} explains position bias due to "causal masking inherently biases attention toward earlier positions, as tokens in deeper layers attend to increasingly contextualized representations of earlier tokens". Our results reveal exceptions where later positions are desired, highlighting the complexity of the problem. \textit{Training data biases}, such as serial-position effects in corpora, shape how models prioritize sequence positions~\citep{wu2025emergence, guo2024serial}. 

% However, the interplay between position bias and multilingual reasoning remains underexplored.  

\textbf{Interplay of Culture, Language, and Model Design.}
\label{sec:cultural_interplay}
The way we perceive the world is influenced not only by our culture~\citep{oyserman2008does} but also by the language we speak~\citep{boroditsky2003sex}. The latter point is particularly relevant for  LLMs, since they are trained on specific languages. Recent studies have shown that multilingual LLMs often initiate their "thinking process" in English, pivoting to the prompt's original language in the middle layers~\citep{zhang2024found, peysakhovich2023attention}. These models exhibit lower lexical naturalness in non-English languages, with the naturalness gap being more pronounced for languages structurally distant from English~\citep{guo2024large}. While the volume of training data plays a crucial role~\citep{arnett2024language}, \textit{linguistic complexity --- including lexical and morphological variations across languages --- must also be considered}~\citep{ghosh2024morphology, dang2024morphology, ismayilzada2024evaluating}. Additionally, architectural design choices affect languages in different ways; for instance, removing positional encoding from language models would most degrade performance in languages with limited morphological systems~\citep{ghosh2024morphology}.
%\footnote{This raises an important question: Has the development of LLM architectures over the last decade been primarily geared toward English? As noted, "English-centric practices in NLP may have impeded progress for language models in other languages"~\citep{arnett2024language}. While this issue is important, it falls outside the scope of this paper}.

At the same time, most of \textbf{bias mitigation approaches} evaluate their performance in English~\citep{zhang2024found,peysakhovich2023attention, zhang2024can, yu2024mitigate,wang2024eliminating}. These approaches fall into two categories: \textbf{prompt-based techniques} and \textbf{architectural interventions}. \textbf{Architectural methods}---such as positional encodings, alternative masking schemes, and calibration mechanisms---address root causes but often require retraining and introduce computational overhead~\citep{zhang2024found, wu2025emergence, zhao2021calibrate}. \textbf{Prompt-based strategies}, including query-aware contextualization and recency prompting, aim to redirect attention dynamically~\citep{peysakhovich2023attention, wang2024eliminating, yu2024mitigate}. 
% However, their effectiveness is inconsistent and task-dependent~\citep{guo2024serial}. 
We focus on prompt-based strategies, using as a starting point the work done by~\citep{zhang2024can}. The authors studied whether a model can improve its performance when given explicit placement of the correct answer. They used two types of instructions --- \textit{relative} and \textit{absolute} --- and found that models lack relative awareness and that implicit information about absolute placement of the correct prompt marginally improves model performance.  The main difference with this approach and ours is that authors in~~\citep{zhang2024can} use relevant distractors, while in our work we use random distractors.
%However, the authors did not study a scenario when the placement prompt is absent at all. 

\label{sec:rag_reorder}
\textbf{Practical considerations.} Position bias affects Chain of Thought Strategies (CoT). CoT struggles with position bias even when reasoning steps are correct, models often fail to retrieve evidence from middle positions~\citep{zhang2024found}.  In \citep{zhang2024found, yu2024mitigate}, authors analyzed error propagation in multi-hop reasoning.  In RAG systems, one of the mitigating strategies is context ordering
~\citep{wang2024rear, alessio2024improving, jin2024long}
 (we discuss these approaches in Appendix~\ref{app:rag}). While conventional approaches often assume a monotonic relationship between document position and attention (e.g., prioritizing the first/last positions), our analysis reveals that position bias may exhibit language-specific patterns and is not always maximized at early tokens. This observation challenges assumptions in methods like \textbf{Long-Context RAG}, which rely on fixed position prioritization, and highlights the need for language-adaptive reordering.

While predictive entropy is widely used to quantify model uncertainty~\citep{huang2024survey, sychev2025llm}, its relationship with position bias remains unexplored. \citep{duan2023shifting} note that uncertainty estimates can be token-biased, but how position bias interacts with uncertainty dynamics is unclear.  

\section{Methods}
\subsection{Position bias formalization} 
To evaluate the effect of position bias, we consider a question-answering task. For each question $Q$, we assume the existence of a ground truth answer $A$ and a set of $N$ contexts $\text{ctx}_1, \dots, \text{ctx}_N$. In our experimental setup, exactly one context is relevant to each question, while the remaining $N-1$ contexts are randomly sampled from the dataset. The response of the model, denoted as $\text{Model}$, using a prompt function $\text{Prompt}$, to the question $Q$ when the relevant context is placed at position $i$ is given by: $R_i = \text{Model}(\text{Prompt}(Q, \text{ctx}_1, \dots, \text{ctx}_N, i))$. We define that the model exhibits position bias on dataset $D$ toward position $i$ over position $j$ if the expected \textit{Accuracy} when the relevant context is at position $i$ exceeds that at position $j$:

$$\mathop{\mathbb{E}}_{(Q, A)\sim D} \left[ \text{Acc}(A, R_i) \right] > \mathop{\mathbb{E}}_{(Q, A)\sim D} \left[ \text{Acc}(A, R_j) \right]$$

\subsection{Context Placement Strategies}\label{sec:context_placement}

\begin{figure*}[htbp]
\centerline{\includegraphics[width=14.5cm]{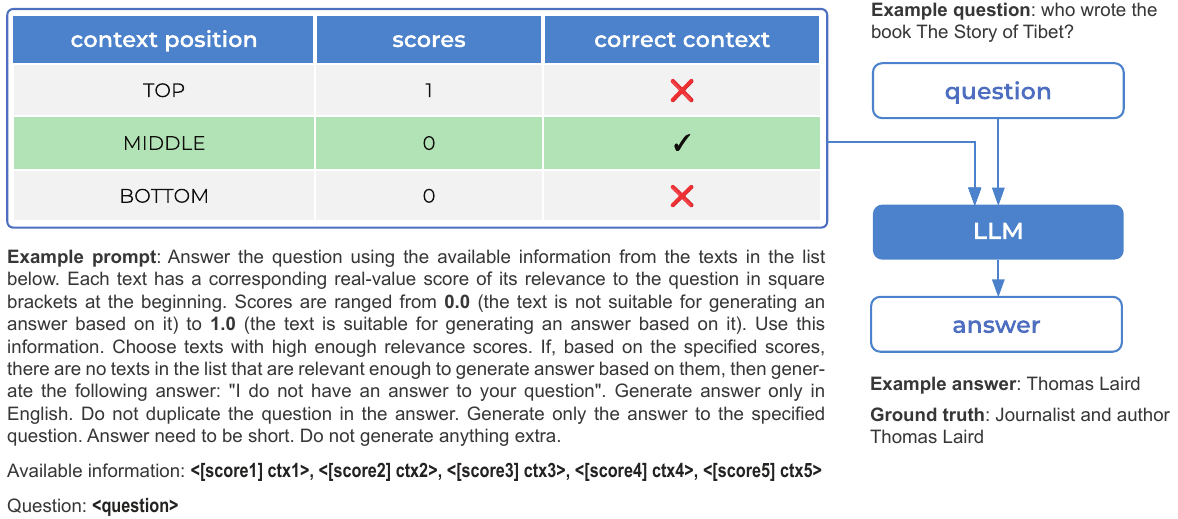}}
\caption{Experiment Structure. For each question, the relevant context is placed in one of three positions: top, middle, or bottom of the context list. Each context is assigned a binary score (0 or 1), indicating its relevance to the question. Three scoring strategies are evaluated: \textbf{Aligned}, where the relevant context receives a score of 1 and all others are assigned 0; \textbf{All Zero}, in which every context, including the relevant one, is scored 0; \textbf{No Scores}, where relevance scores are entirely omitted. This experimental design assesses the influence of scoring mechanisms and answer positioning on model performance under varying degrees of contextual guidance.}
\label{fig:experiment_structure}
\end{figure*}

This experimental series examines whether explicit information regarding the relevance of contexts to a given query can improve performance. We investigate a practical scenario in which context relevance can be quantified (e.g., through cosine similarity between context and question embeddings) and assess the impact of integrating relevance scores into prompts on context selection. The experimental framework is illustrated in Figure~\ref{fig:experiment_structure}.

\noindent\textbf{Scoring Strategies.} Three distinct scoring strategies are evaluated. The \textbf{Aligned} strategy assigns a relevance score of $rs=1$ to contextually relevant information and $rs=0$ to all other contexts. The \textbf{All Zero} strategy assigns $rs=0$ to all contexts, including those that are relevant, to test the hypothesis that intentional mislabeling will degrade model performance if relevance scores influence context selection. The \textbf{No Scores} strategy omits relevance scores entirely from the prompts to assess whether explicit scoring contributes to model efficacy. In the experimental setup, relevance scores $rs_1, \ldots, rs_N$ were incorporated into the prompt where applicable. As illustrated in Figure~\ref{fig:experiment_structure}, three positions for relevant contexts are considered: TOP (first), MIDDLE ($N/2$), and BOTTOM (last).

\noindent\textbf{Context Volume.} Prior research has shown that longer contexts can exacerbate position bias \citep{baker2024lost, peysakhovich2023attention}. In our work, we default to \textbf{five} contexts based on preliminary experiments. We vary the number of contexts, setting \( N \in \{5, 10, 15\} \) , and find that while increasing N  does degrade performance for some models, the effect is relatively modest. Given limited computational resources, we therefore opted to use five contexts in our main experiments. Preliminary results are provided in Appendix~\ref{app:number_of_prompts}.

\subsection{Average Predictive Entropy}
We adopt the \textit{token-wise entropy} framework introduced by~\citep{lu2021fantastically}, normalized by the total number of tokens, to quantify uncertainty in model responses~\citep{lu2021fantastically, wang2025word}. Let \( x \) represent the input prompt and \( s = \{z_1, z_2, \dots, z_n\} \) denote a generated completion sequence of \( n \) tokens. For a given large language model (LLM), the conditional probability of generating the \( i \)-th token \( z_i \), given the preceding tokens \( s_{<i} = \{z_1, \dots, z_{i-1}\} \) and the prompt \( x \), is denoted as \( p(z_i \mid s_{<i}, x) \) for \( 1 \leq i \leq n \). The \textit{average predictive entropy}\(^1\) (denoted as \( \textit{PE}_{\text{avg}} \)) is defined as:

\begin{small}
\begin{equation}\label{eq:PE_avg}
    \textit{PE}_{\text{avg}}(s, x) = -\frac{1}{n} \log p(s \mid x) = \frac{1}{n} \sum_{i=1}^n -\log p(z_i \mid s_{<i}, x).
\end{equation}  
\end{small}

This formulation computes the average uncertainty per token by decomposing the joint probability \( p(s \mid x) \) into a product of conditional probabilities using the chain rule. The normalization by \( n \) ensures comparability across sequences of varying lengths, consistent with the interpretation of entropy as a measure of "average uncertainty".

\section{Experiment Set-Up}
\subsection{Datasets}
In this study, we employed three open-ended question-answering datasets encompassing five languages characterized by divergent syntactic structures and semantic distributions. The statistics for these datasets are summarized in Table~\ref{tab:datasets}, with a comprehensive description provided in Appendix~\ref{app:datasets}. These datasets were selected for two principal reasons: (1) they are well-established within the research community, ensuring familiarity and reproducibility; (2) the context lengths for question-answer pairs are sufficiently concise (less than 4096 characters), allowing multiple instances to be included in a single prompt without exceeding the maximum sequence length constraints of the language model.

\begin{table}
\centering
\small
\begin{tabular}{lcc}
\toprule
\textbf{Language} & \textbf{Source} & \textbf{Size} \\
\midrule
English & SQuAD2.0~\citep{rajpurkar2018know} & 150k  \\
%\hline
Russian & MTS-SQuAD (\href{https://huggingface.co/datasets/MTS-AI-SearchSkill/MTSBerquad}{link}) & 60k  \\
%\hline
German & MLQA~\citep{lewis2020mlqa} & ~5k  \\
%\hline
Hindi & MLQA~\citep{lewis2020mlqa} & ~5k  \\
%\hline
Vietnamese & MLQA~\citep{lewis2020mlqa} & ~5k  \\
\bottomrule
\end{tabular}
\caption{Summary of datasets utilized in this study, categorized by language, with corresponding sources and associated question-answer pairs sizes}
\label{tab:datasets}
\end{table}

\textbf{Preprocessing.} To accommodate computational constraints, the analysis was limited to 2,000 question-answer (QA) pairs per language. A two-stage preprocessing pipeline was applied prior to sampling to ensure data quality and consistency: (1) duplicate removal, in which all redundant QA pairs were excluded; (2) answer validation, where pairs lacking valid responses, such as those with missing or ambiguous answers, were discarded.

\subsection{Models}
\label{sec:models}
To investigate whether position bias in large language models (LLMs) arises from model-specific design and training choices or from language-specific characteristics, we evaluate five popular open-source multilingual models: \textbf{Qwen2.5-7B-Instruct}\citep{qwen2025qwen25technicalreport}, \textbf{Llama3-8B-Instruct}\citep{touvron2023llamaopenefficientfoundation}, \textbf{DeepSeek-7B-Chat}\citep{deepseek-llm}, \textbf{Gemma-7B-it}\citep{gemmateam2024gemmaopenmodelsbased}, and \textbf{Mistral-7B-Instruct-v0.3}\citep{jiang2023mistral7b}. All models support English, German, Russian, Vietnamese, and Hindi, yet differ in architecture and training paradigms. This selection enables us to disentangle model driven factors of position bias such as attention mechanisms and training data composition from language-related influences.   It is important to note that \textbf{DeepSeek-7B-Chat} performed poorly on Hindi, therefore, we excluded its Hindi results from our analysis. 
\section{Evaluation}
\subsection{LLM as a Judge}
Traditional statistical evaluation metrics of open-ended generations, such as \textit{BLEU} \citep{papineni2002bleu}, \textit{ROUGE} \citep{lin2004rouge}, and \textit{Meteor Universal} \citep{denkowski2014meteor}, are limited in their ability to differentiate between syntactically similar yet semantically distinct texts. Although semantic evaluation methods like \textit{BERTScore} \citep{zhang2019bertscore} were developed to overcome these shortcomings, our experimental findings indicate that BERTScore exhibits insufficient discriminative power, frequently failing to capture subtle distinctions between correct and incorrect responses. Consequently, we employ the \textit{LLM as a judge} framework \citep{zheng2023judging} and select \textbf{Mistral Large}\footnote{https://mistral.ai/news/mistral-large-2407} as the evaluator for the following reasons: (1) prior research demonstrates its strong alignment with human judgments and generalizability across diverse tasks \citep{bavaresco2406llms, kim2024prometheus}; (2) it provides a freely accessible API for research, facilitating large-scale evaluation; (3) its architectural design differs from the majority of models used for response generation, thereby reducing potential bias toward self-generated outputs. The evaluator assesses question-answer pairs using a structured prompt that includes the question, ground truth, and model-generated answer. It assigns a label of $1$ for correct answers and $0$ for incorrect ones; accuracy is adopted as the primary metric. Further details regarding the prompts and evaluation methodology are provided in Appendix~\ref{app:lmjudge}.

\subsection{Human Evaluation}
To validate the reliability of the large language model (LLM) as an evaluative judge, human annotation was conducted on a set of 150 questions in both English and Russian. The responses generated by the Llama model were annotated using the \textit{Overlap-3} metric, with domain experts adhering to the same evaluation criteria as the automated judge. Inter-annotator agreement was quantified using Krippendorff’s $\alpha$ \citep{krippendorff2011computing}, yielding a mean $\alpha = 0.755$, which indicates a high degree of assessment reliability. Further evaluation of  alignment between the automated judge and human annotators is conducted by computing the Pearson correlation coefficient \( r \) between the judge’s scores and the majority vote derived from human annotations. A strong mean correlation of \( r = 0.716 \) was observed, indicating substantial agreement. Additional details regarding the annotation procedure are provided in Appendix \ref{app:lmjudge}.

\section{Experiments and Results}
\label{sec:results}
% During the analysis of statistical significance, we applied pairwise t-tests with the Holm correction of three combinations: TOP vs. MIDDLE, TOP vs. BOTTOM and MIDDLE vs. BOTTOM. A language setup is considered to have a significant effect if at least two of the three comparisons show a statistically significant difference. According to the results, two languages with a significant effect were identified for the Llama and Qwen models, and it is noteworthy that in both cases they are the same languages: German and Vietnamese.

\subsection{Sanity Check for evaluation procedure}
\begin{figure}[ht!]
\centering
\centerline{\includegraphics[width=8cm]{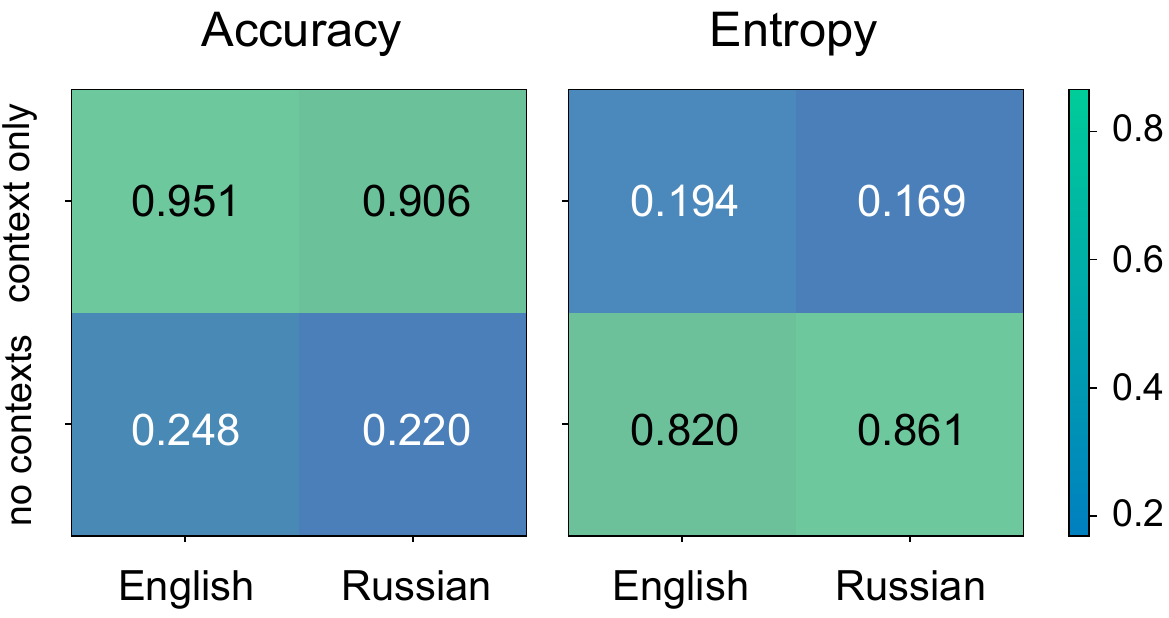}}
\caption{Results in the presence and absence of contextual information for Llama3.1 8B model on SQuAD2.0 (en) and MTS-SQuAD (ru). We use a single relevant context to evaluate whether the provision of such information enhances model performance or if the model can generate accurate responses based solely on its internal knowledge.}
\label{fig:sanity_check}
\end{figure}

Before proceeding with the remaining experiments, we first perform a simple sanity check to demonstrate that without any relevant context model performance drops significantly on two benchmarks: SQuAD2.0 (English) and MTS-SQuAD (Russian) with Llama3.1 8B model. In these experiment we wanted to answer a question if the model can generate accurate responses
based solely on its internal knowledge. Additionally, we measure predictive entropy and show that it increases when no relevant context is provided. While these results are neither novel nor surprising, they serve to validate our overall evaluation procedure. The results are presented in Figure~\ref{fig:sanity_check}.

\subsection{Overall Results Across Languages and Models}
The second step of our evaluation assesses overall model performance and dataset difficulty. Tables~\ref{tab:model_difficulty} and~\ref{tab:dataset_difficulty} show performance averaged across all positional configurations and prompting strategies (full results are provided in Appendix~\ref{app:extres}). \textbf{Language performance} is highest for English (SQuAD2.0: 0.896) and Russian (MT-SQuAD: 0.865), but drops for German (0.628) and especially Hindi and Vietnamese ($\approx$0.47), indicating greater difficulty in lower-resource or morphologically complex languages.  \textbf{For model performance}, Qwen2.5-7B leads (0.754), followed by Mistral-7B (0.743), while DeepSeek-7B (0.711) and Llama3.1-8B (0.690) perform moderately. Gemma-7B lags substantially (0.486), suggesting weaker multilingual capabilities or architectural limitations.

\begin{table}[h!]
\centering
\begin{minipage}{0.48\linewidth}
\centering
\renewcommand{\arraystretch}{1.25}
\small
\setlength{\tabcolsep}{4pt}
\begin{tabular}{cc}
\hline
\textbf{Dataset} & \textbf{Acc} \\
\hline
SQuAD2.0 (en) & 89.6 \\
MTS-SQuAD (ru) & 86.5 \\
MLQA (de) & 62.8 \\
MLQA (hi) & 47.1 \\
MLQA (vi) & 47.5 \\
\hline
\end{tabular}
\caption{\textbf{Performance on specific language} averaged by models, positions and strategies}
\label{tab:model_difficulty}
\end{minipage}\hfill
\begin{minipage}{0.48\linewidth}
\centering
\renewcommand{\arraystretch}{1.25}
\small
\setlength{\tabcolsep}{4pt}
\begin{tabular}{cc}
\hline
\textbf{Model} & \textbf{Acc} \\
\hline
DeepSeek 7B & 71.1 \\
Gemma 7B & 48.6 \\
Mistral 7B & 74.3 \\
Llama3.1 8B & 69.0 \\
Qwen2.5 7B & 75.4 \\
\hline
\end{tabular}
\caption{\textbf{Performance of specific model} averaged by datasets, positions, and strategies}
\label{tab:dataset_difficulty}
\end{minipage}
\end{table}

\subsection{Position bias is mostly driven by models}
While the “Lost in the Middle” phenomenon holds true, all models perform worst in the middle, as shown in Table~\ref{tab:middle}, preferences for the beginning versus the end vary across datasets and models.   We observe both model-specific bias (Table~\ref{tab:model_bias}) and language-specific bias (Table~\ref{tab:lang_bias}). However, the aggregated results in Table~\ref{tab:bias_total} suggest that positional bias is primarily model-driven, as discrepancies across positions are more pronounced between models than across languages. Specifically, DeepSeek-7B, Mistral-7B, and Qwen2.5-7B exhibit a preference for later positions, whereas Llama3.1-8B and Gemma-7B favor earlier ones. The strongest positional effect is observed for DeepSeek-7B and Llama3.1-8B, with a performance difference of approximately 0.1 points between positions. 

% =============================

\begin{table}
\centering
\small
\begin{tabular}{cp{1cm}p{1cm}|ccc}
\toprule
\multicolumn{3}{c|}{\textbf{Strategy}} & \multicolumn{3}{c}{\textbf{Position}} \\
\cmidrule(r){1-3} \cmidrule(l){4-6}
Aligned & All Zero & No Scores & TOP & MID & BOT \\
\midrule
0.695 & \cellcolor{blue!30}0.610 & 0.721 & 0.684 & \cellcolor{blue!30}0.666 & 0.676 \\
\bottomrule
\end{tabular}
\caption{Accuracy score for different \textbf{Strategies} of position-bias elimination(Aligned, All Zero, No Scores) and \textbf{Positions} of relevant context(TOP, MIDDLE, BOTTOM). The results in each column are averaged across all other setups. The worst performance is highlighted in \textbf{blue}.}
\label{tab:middle}
\end{table}

\begin{table}[ht]
\centering
\small % Reduce font size slightly
\begin{tabular}{ c | c c | c c }
%\cmidrule(r){1-2} \cmidrule(l){4-5}
\toprule
%\hhline{|-|-|-|-|-|}
\textbf{Position} & \textbf{Dataset} & \textbf{Acc} & \textbf{Model} & \textbf{Acc} \\
%\hhline{|-|-|-|-|-|}
\hline
TOP    & \multirow{3}{0.5cm}{en} & \cellcolor{green!10} \textbf{0.904} & \multirow{3}{1.8cm}{DeepSeek 7B} & 0.683 \\
MID &                               & 0.894 &                                     & 0.697 \\
BOT &                               & 0.891 &                                     & \cellcolor{green!10}\textbf{0.753} \\
\hhline{|-|-|-|-|-|}
TOP    & \multirow{3}{0.5cm}{ru} & 0.861 & \multirow{3}{1.8cm}{Gemma 7B}         & \cellcolor{green!10}\textbf{0.497} \\
MID &                               & 0.861 &                                     & 0.492 \\
BOT &                               & \cellcolor{green!10}\textbf{0.873} &                                     & 0.470 \\
\hhline{|-|-|-|-|-|}
TOP    & \multirow{3}{0.5cm}{de}    & \cellcolor{green!10}\textbf{0.646} & \multirow{3}{1.8cm}{Mistral 7B}       & 0.748 \\
MID &                               & 0.615 &                                     & 0.729 \\
BOT &                               & 0.624 &                                     & \cellcolor{green!10}\textbf{0.751} \\
\hhline{|-|-|-|-|-|}
TOP    & \multirow{3}{0.5cm}{hi}    & \cellcolor{green!10}\textbf{0.493} & \multirow{3}{1.8cm}{Llama3.1 8B}      & \cellcolor{green!10}\textbf{0.743} \\
MID &                               & 0.452 &                                     & 0.671 \\
BOT &                               & 0.469 &                                     & 0.655 \\
\hhline{|-|-|-|-|-|}
TOP    & \multirow{3}{0.5cm}{vi}    & 0.480 & \multirow{3}{1.8cm}{Qwen2.5-7B}       & 0.752 \\
MID &                               & 0.463 &                                     & 0.744 \\
BOT &                               & \cellcolor{green!10}\textbf{0.483} &                                     & \cellcolor{green!10}\textbf{0.767} \\
%\hhline{|-|-|-|-|-|}
\bottomrule
\end{tabular}
\caption{Model-wise and Language-wise position bias, highest accuracies reflect bias and are highlighted in \textbf{bold-green}. Results represent accuracy averaged across all  strategies \textbf{Aligned}, \textbf{All Zero} and \textbf{No Scores} presented in Table~\ref{tab:lang_bias} and Table~\ref{tab:model_bias}. Position indicates correct context position among all contexts.}
\label{tab:bias_total}
\end{table}

% From Table~\ref{tab:metrics_combined_main}, it is observed that Qwen generally achieves the highest performance when the correct context is placed at the bottom, whereas Llama excels when the correct context is positioned at the top across all languages. Notably, Qwen demonstrates superior performance for Vietnamese and Russian in specific scenarios. However, despite its overall stronger average performance, Llama outperforms Qwen in Vietnamese. This discrepancy may be attributed to an alignment between Vietnamese language-specific position bias and the inherent position bias of Llama. Overall, our findings suggest that while language-specific position biases may exist, they are likely overshadowed by the dominant positional preferences intrinsic to the models themselves. These results challenge conventional assumptions regarding primacy effects in large language models.

% ----------------------------

\begin{table}[ht!]
\centering
\label{tab:combined}

\begin{minipage}[t]{0.48\textwidth}
\centering
\resizebox{\textwidth}{!}{%
\renewcommand{\arraystretch}{1.3}
\begin{tabular}{ccccc}
\hhline{|-|-|-|-|-|}
\textbf{Dataset} & \textbf{Position} & \textbf{Aligned} & \textbf{All Zero} & \textbf{No Scores} \\ 
\hhline{|-|-|-|-|-|}
\multirow{3}{*}{SQuAD2.0 (en)}
  & TOP    & \colorcell{0.918} & \colorcell{0.874} & \colorcell{0.921} \\ %\cline{2-5}
  & MID & \colorcell{0.919} & \colorcell{0.844} & \colorcell{0.920} \\ %\cline{2-5}
  & BOT & \colorcell{0.916} & \colorcell{0.835} & \colorcell{0.922} \\ \hhline{|-|-|-|-|-|}
\multirow{3}{*}{MTS-SQuAD (ru)}
  & TOP    & \colorcell{0.877} & \colorcell{0.823} & \colorcell{0.883} \\ %\cline{2-5}
  & MID & \colorcell{0.873} & \colorcell{0.825} & \colorcell{0.883} \\ %\cline{2-5}
  & BOT & \colorcell{0.889} & \colorcell{0.841} & \colorcell{0.888} \\ \hhline{|-|-|-|-|-|}
\multirow{3}{*}{MLQA (de)}
  & TOP    & \colorcell{0.674} & \colorcell{0.579} & \colorcell{0.684} \\ %\cline{2-5}
  & MID & \colorcell{0.658} & \colorcell{0.525} & \colorcell{0.661} \\ %\cline{2-5}
  & BOT & \colorcell{0.664} & \colorcell{0.542} & \colorcell{0.666} \\ \hhline{|-|-|-|-|-|}
\multirow{3}{*}{MLQA (hi)}
  & TOP    & \colorcell{0.518} & \colorcell{0.415} & \colorcell{0.544} \\ %\cline{2-5}
  & MID & \colorcell{0.483} & \colorcell{0.353} & \colorcell{0.520} \\ %\cline{2-5}
  & BOT & \colorcell{0.519} & \colorcell{0.363} & \colorcell{0.525} \\ \hhline{|-|-|-|-|-|}
\multirow{3}{*}{MLQA (vi)}
  & TOP    & \colorcell{0.466} & \colorcell{0.410} & \colorcell{0.565} \\ %\cline{2-5}
  & MID & \colorcell{0.452} & \colorcell{0.379} & \colorcell{0.558} \\ %\cline{2-5}
  & BOT & \colorcell{0.486} & \colorcell{0.397} & \colorcell{0.566} \\ \hhline{|-|-|-|-|-|}
\end{tabular}}
\captionof{table}{\textbf{Language specific bias} with resulting accuracy averaged by models.}
\label{tab:lang_bias}
\end{minipage}
\hfill
\vspace{1cm}
\begin{minipage}[t]{0.48\textwidth}
\centering
\resizebox{\textwidth}{!}{%
\renewcommand{\arraystretch}{1.3}
\begin{tabular}{ccccc}
\hline
\textbf{Model} & \textbf{Position} & \textbf{Aligned} & \textbf{All Zero} & \textbf{No Scores} \\
\hline
\multirow{3}{*}{DeepSeek 7B}
  & TOP    & \colorcell{0.683} & \colorcell{0.669} & \colorcell{0.696} \\ %\cline{2-5}
  & MID & \colorcell{0.702} & \colorcell{0.684} & \colorcell{0.704} \\ %\cline{2-5}
  & BOT & \colorcell{0.761} & \colorcell{0.743} & \colorcell{0.756} \\ \hline
\multirow{3}{*}{Gemma 7B}
  & TOP    & \colorcell{0.507} & \colorcell{0.507} & \colorcell{0.573} \\ %\cline{2-5}
  & MID & \colorcell{0.515} & \colorcell{0.515} & \colorcell{0.571} \\ %\cline{2-5}
  & BOT & \colorcell{0.485} & \colorcell{0.485} & \colorcell{0.556} \\ \hline
\multirow{3}{*}{Mistral 7B}
  & TOP    & \colorcell{0.755} & \colorcell{0.709} & \colorcell{0.779} \\ %\cline{2-5}
  & MID & \colorcell{0.743} & \colorcell{0.684} & \colorcell{0.761} \\ %\cline{2-5}
  & BOT & \colorcell{0.762} & \colorcell{0.716} & \colorcell{0.776} \\ \hline
\multirow{3}{*}{Llama3.1 8B}
  & TOP    & \colorcell{0.772} & \colorcell{0.648} & \colorcell{0.808} \\ %\cline{2-5}
  & MID & \colorcell{0.720} & \colorcell{0.517} & \colorcell{0.777} \\ %\cline{2-5}
  & BOT & \colorcell{0.735} & \colorcell{0.481} & \colorcell{0.748} \\ \hline
\multirow{3}{*}{Qwen2.5 7B}
  & TOP    & \colorcell{0.768} & \colorcell{0.715} & \colorcell{0.772} \\ %\cline{2-5}
  & MID & \colorcell{0.750} & \colorcell{0.715} & \colorcell{0.767} \\ %\cline{2-5}
  & BOT & \colorcell{0.780} & \colorcell{0.745} & \colorcell{0.777} \\ \hline
\end{tabular}}
\captionof{table}{\textbf{Model specific bias} with resulting accuracy averaged by datasets.}
\label{tab:model_bias}
\end{minipage}

\end{table}

\subsection{Positional guidance}
\textbf{Sensitivity to prompt guidance.} For all considered languages incorrect relevance scoring leads to significant performance decrease. It varies from $4.3\%$ drop for English (0.918 \textit{Aligned} vs 0.874 \textit{All Zero}) to $15.6\%$ for Hindi (0.519 \textit{Aligned} vs 0.363 \textit{All Zero}). All models except Gemma exhibit pronounced sensitivity to positional cues when contextual scoring is perturbed. The introduction of misleading scores results in a marked decline in accuracy ranging from $1.4\%$ drop for DeepSeek performance (from $0.683$ \textit{Aligned} to $0.669$ \textit{All-Zero}), to $25.4\%$ drop for Llama performance (from $0.735$ \textit{Aligned} to $0.481$ \textit{All-Zero}). The absence of fluctuations in accuracy for Gemma is probably observed, because of poor quality of this model for all setups and general inability to handle positional guidance. Overall, the majority of setups are sensitive to positional guidance, so such techniques could be utilized to handle position bias.

\textbf{Score Omission Enhances Robustness.} Notably, the \textit{No Scores} consistently outperforms other strategies. For languages there is a little improvement for Russian in BOTTOM position( $0.889$ \textit{Aligned} vs $0.888$ \textit{No Scores}). Similarly, only for Qwen2.5 7B( $0.780$ \textit{Aligned} vs $0.777$ \textit{No Scores}) and DeepSeek 7B ($0.761$ \textit{Aligned} vs $0.756$ \textit{No Scores}) in BOTTOM position the minor accuracy improvements are observed. For all other setups the performance is deteriorate, ranging from $0.1\%$ to $10.2\%$ accuracy drop for languages and from $0.2\%$ to $7.1\%$ for models (\textit{Aligned} vs \textit{No Scores}). This performance degradation is particularly pronounced in low-resource languages such as Vietnamese. These findings challenge previous works~\citep{zhang2024can} and force to thorough validation of guiding strategies.

% Similarly, Qwen’s preference for BOTTOM-placed contexts does not yield consistent accuracy gains (e.g., Vietnamese MLQA: $0.718$ [TOP] vs. $0.726$ [BOTTOM]). This disconnect between positional guidance and performance improvement highlights a critical limitation: models follow position biases regardless of their relevance to answer correctness. In practical RAG systems, such behavior risks amplifying retrieval errors or misaligned document rankings, rendering positional optimization ineffective for enhancing downstream task outcomes.   

\textbf{Language-Specific Sensitivity.} High-resource languages, such as English and Russian, demonstrated minimal performance variation across scenarios ($\Delta < 2.5\%$ in English), whereas languages on \textit{MLQA} benchmark exhibited more pronounced differences. A phenomenon potentially attributable to orthographic or syntactic properties that may mediate position bias.

\subsection{Highest Entropy is not associated with the lowest performance.}
\label{res:entropy}
While we observe that models perform most poorly in the middle, the highest predictive entropy is not always associated with the middle position (see Table~\ref{tab:entropy}). We speculate that this phenomenon may arise due to token homogenization and provide a formal analysis in Appendix~\ref{app:homogenization}.

\begin{table}[ht]
\centering
\small
\begin{tabular}{p{1.1cm}p{1.0cm}|p{0.75cm}p{0.75cm}p{0.75cm}p{0.75cm}}
\toprule
% \multicolumn{2}{c}{Aggregation by Dataset} \\
% \cmidrule(r){1-2}
\textbf{Model} & \textbf{Position} & \textbf{Aligned} & \textbf{All Zero} & \textbf{No Scores} & \textbf{Mean} \\
   \hhline{|-|-|-|-|-|-|}
\multirow{3}{1.1cm}{DeepSeek 7B} & TOP    & \colorcellentropy{0.257} & \colorcellentropy{0.258} & \colorcellentropy{0.233} & \cellcolor{blue!40} 0.250 \\
                                  & MID & \colorcellentropy{0.250} & \colorcellentropy{0.254} & \colorcellentropy{0.227} & \colorcellentropy{0.243} \\
                                  & BOT & \colorcellentropy{0.233} & \colorcellentropy{0.238} & \colorcellentropy{0.213} & \colorcellentropy{0.228} \\
\hline
\multirow{3}{1.1cm}{Gemma 7B}         & TOP    & \colorcellentropy{0.189} & \colorcellentropy{0.189} & \colorcellentropy{0.194} & \colorcellentropy{0.191} \\
                                  & MID & \colorcellentropy{0.187} & \colorcellentropy{0.187} & \colorcellentropy{0.196} & \colorcellentropy{0.190} \\
                                  & BOT & \colorcellentropy{0.189} & \colorcellentropy{0.189} & \colorcellentropy{0.200} & \cellcolor{blue!40} 0.193 \\
\hline
\multirow{3}{1.1cm}{Mistral 7B}       & TOP    & \colorcellentropy{0.194} & \colorcellentropy{0.194} & \colorcellentropy{0.217} & \colorcellentropy{0.202} \\
                                  & MID & \colorcellentropy{0.202} & \colorcellentropy{0.205} & \colorcellentropy{0.221} & \cellcolor{blue!40} 0.209 \\
                                  & BOT & \colorcellentropy{0.199} & \colorcellentropy{0.201} & \colorcellentropy{0.215} & \colorcellentropy{0.205} \\
\hline
\multirow{3}{1.1cm}{Llama3.1 8B}      & TOP    & \colorcellentropy{0.248} & \colorcellentropy{0.258} & \colorcellentropy{0.217} & \colorcellentropy{0.241} \\
                                  & MID & \colorcellentropy{0.251} & \colorcellentropy{0.241} & \colorcellentropy{0.232} & \colorcellentropy{0.241} \\
                                  & BOT & \colorcellentropy{0.254} & \colorcellentropy{0.238} & \colorcellentropy{0.241} &  \cellcolor{blue!40} 0.244 \\
\hline
\multirow{3}{1.1cm}{Qwen2.5 7B}       & TOP    & \colorcellentropy{0.101} & \colorcellentropy{0.097} & \colorcellentropy{0.105} & \colorcellentropy{0.101} \\
                                  & MID & \colorcellentropy{0.106} &  \colorcellentropy{0.104} & \colorcellentropy{0.112} &  \cellcolor{blue!40} 0.107 \\
                                  & BOT & \colorcellentropy{0.106} & \colorcellentropy{0.101} & \colorcellentropy{0.108} & \colorcellentropy{0.105} \\
   \hhline{|-|-|-|-|-|-|}
\end{tabular}
\caption{Predictive Entropy - PE across positions, models and strategies. Highest mean values across strategies are highlighted in purple.}
\label{tab:entropy}
\end{table}

\subsection{Word Order Analysis}
We additionally perform investigation if there is the relationship between the position
of relevant context, model behavior, and the dominant word order of a language. Since we find no evidence to suggest that position bias influences models to favor specific word orders we discuss our methodology and detailed results in Appendix~\ref{app:wordorder}.

\section{Conclusion}

This study reveals that position bias in multilingual LLMs is primarily model-driven, contradicting the assumed universal early-token preference—with architectures like Qwen2.5-7B and DeepSeek-7B favoring late positions, while Llama-3.1-8B prefers early ones. Language-specific effects exist but are secondary. Surprisingly, explicitly signaling context relevance via prompt-based relevance scores consistently harms performance, especially in low-resource languages. Moreover, the worst accuracy (when relevant context is in the middle) does not correspond to higher output entropy, indicating models are confidently wrong under positional disadvantage. These findings challenge common RAG and prompting practices and underscore the need for model- and language-aware context handling.
\section{Limitations}
\label{sec:limitations}
Since LLM-as-a-Judge was utilized to assess the correctness of open-ended question-answering task, our methodology depends on its performance critically.

Our evaluation used 2,000 question–answer pairs per language. Across nine experimental conditions and five models, this amounts to $9\times5\times2000=70 000$ model evaluations per language, a computationally intensive effort. Given this scale, we took care to ensure statistical rigor. Specifically, we performed pairwise t-tests with Holm–Bonferroni correction for the three positional comparisons: (1) top vs. middle, (2) top vs. bottom, and (3) middle vs. bottom. For every dataset and model, at least one of these (two on average) comparisons yielded statistically significant differences in accuracy (p < 0.05). However, when aggregating results across models, we found no significant differences between languages. The same pattern held for predictive entropy values.

% \textbf{Word Order Analysis.} We recognize that a rigorous analysis of subject-verb-object relationships would require examining a broader range of syntactic dependencies. However, such detailed linguistic inquiry lies beyond the methodological scope of this study.

% \textbf{Entropy Analysis.} Our attention entropy analysis is subject to two principal limitations. First, although token homogenization (the assumption that all tokens are treated uniformly) warrants deeper mechanistic investigation, such an exploration lies beyond the scope of this study. Second, we have not formally established a theoretical connection between attention entropy and predictive entropy. This gap precludes a comprehensive validation of our hypothesis that aligning positional and contextual attention patterns influences predictive entropy.

\paragraph{Acknowledgment on LLM assisted writing:}
This paper used open access Qwen3-Max, in some parts of the paper,  for proofreading and text rephrasing in accordance with formal style.

\bibliography{custom}
\section*{Appendix}
\begin{appendix}
\label{sec:appendix}

\section{RAG systems with context reordering}
\label{app:rag}

\textbf{REAR}~\citep{wang2024rear} – integrates document relevance scores into LLMs via embeddings, guiding generation to use internal knowledge (low relevance) or external evidence (high relevance).  
    
\textbf{Long-Context LLMs Meet RAG}~\citep{jin2024long} – addresses the "lost-in-the-middle" problem by reordering retrieved documents, placing highest-scoring ones at sequence boundaries to optimize LLM attention.  
    
\textbf{OP-RAG}~\citep{yu2409defense} – order preserving RAG preserves original document order (vs. sorting chunks), demonstrating improved answer quality through position-aware context organization.  However, authors do not mention multidocument scenario. 
    
\textbf{Clustering \& Reordering RAG}~\citep{alessio2024improving} - cluster sentences by query similarity and sort clusters in descending similarity order for improved retrieval quality.

\section{Datasets}
\label{app:datasets}

\textbf{SQuAD2.0}\footnote{https://huggingface.co/datasets/rajpurkar/squad\_v2}~\citep{rajpurkar2018know} is an English reading-comprehension benchmark built on Wikipedia passages. It combines 100 000 span-answerable questions from SQuAD 1.1~\citep{rajpurkar2016squad} with more than 50 000 adversarial questions whose answers are deliberately absent. 

% The key challenge of  this dataset is ...

\textbf{MTS-SQuAD}\footnote{https://huggingface.co/datasets/MTS-AI-SearchSkill/MTSBerquad} is an extension of the SberQuAD dataset~\citep{efimov2020sberquad} which is a Russian counterpart of SQuAD 2.0. It includes more than 60 000 quesion-answer pairs with improved readability and consistency.

\textbf{MLQA}\footnote{https://github.com/facebookresearch/MLQA}~\citep{lewis2020mlqa} is a multilingual benchmark built from aligned Wikipedia passages including 12 000 question-answer pairs in English and about 5 000 in each of the other six languages: Arabic, German, Spanish, Hindi, Vietnamese, and Simplified Chinese. Among these languages, we utilized German, Hindi and Vietnamese.

\section{Technical Details}
\label{app:techd}

\textbf{Models inference.} To achieve reproducibility of the obtained results, the LLM-inference was performed using a deterministic generation strategy. The following hyperparameter were used/set: "max\_new\_tokens" -- 1024, "do\_sample" -- False, "num\_beams" -- 1.

\textbf{Computational Resources.} The experiments were run in a Docker container on a dedicated server with the following hardware: CPU: AMD Ryzen 9 7900X 12-Core Processor, GPU: NVIDIA GeForce RTX 3090 24GB, RAM: Kingston FURY Beast Black 32GB, SSD: M.2 NVMe Samsung 990 PRO 1T.

\textbf{Required GPU-time for experiments.}  In total it is required approximately $50$ GPU-hours to reproduce the experiments. %In Table \ref{tab:exp_time} you can observe average time that was required to conduct our experiments on GPU.

% \begin{table}[ht!]
% \caption{Averaged GPU-time (in minutes), spent to obtain results for specific experimental setup: dataset, model, and context placement strategy}
% \vspace{1mm}
% \centering
% \renewcommand{\arraystretch}{1.2}
% \begin{tabularx}{\textwidth}{ZZZZZZZ}
% \hline
% \multirow{2}{*}{\textbf{Dataset}} & \multicolumn{3}{c}{\textbf{Qwen}} & \multicolumn{3}{c}{\textbf{Llama}} \\ %\cline{2-7} 
%  & \multicolumn{1}{c}{\textbf{Aligned}} & \multicolumn{1}{c}{\textbf{All Zero}} & \textbf{No Scores} & \multicolumn{1}{c}{\textbf{Aligned}} & \multicolumn{1}{c}{\textbf{All Zero}} & \textbf{No Scores} \\ \hline
% SQuAD2.0 (en) & \multicolumn{1}{c}{15.8} & \multicolumn{1}{c}{15.8} & 14.1 & \multicolumn{1}{c}{17.3} & \multicolumn{1}{c}{17.4} & 14.8 \\ %\hline
% MTS-SQuAD (ru) & \multicolumn{1}{c}{37.7} & \multicolumn{1}{c}{37.3} & 32.7 & \multicolumn{1}{c}{32.7} & \multicolumn{1}{c}{30.7} & 24.8 \\ %\hline
% MLQA (de) & \multicolumn{1}{c}{27.2} & \multicolumn{1}{c}{25.2} & 26.1 & \multicolumn{1}{c}{30.8} & \multicolumn{1}{c}{24.6} & 29.2 \\ %\hline
% MLQA (hi) & \multicolumn{1}{c}{98.1} & \multicolumn{1}{c}{87.3} & 98.6 & \multicolumn{1}{c}{63.1} & \multicolumn{1}{c}{46.1} & 66.9 \\ %\hline
% MLQA (vi) & \multicolumn{1}{c}{33.6} & \multicolumn{1}{c}{30.5} & 32.8 & \multicolumn{1}{c}{40.1} & \multicolumn{1}{c}{35.8} & 30.0 \\ \hline
% \end{tabularx}%
% \label{tab:exp_time}
% \end{table}

\section{Effect of position bias on Entropy}
\label{app:homogenization}

This section formalizes the propagation of position bias toward initial tokens across layers in transformer-based large language models (LLMs) and examines its effect on attention entropy. Under the assumption of a standard multi-head self-attention architecture, we derive conditions under which such bias leads to homogenization of token representations, consequently increasing entropy in the final layer.

\textbf{Notation.}  Let \( X^{(0)} = [x_1, x_2, \dots, x_n] \in \mathbb{R}^{d \times n} \) denote the input token embeddings at layer \( 0 \), where \( x_1 \) is the first token and \( d \) is the embedding dimension. At each layer \( l \geq 1 \), the self-attention operation computes:

\begin{small}
\begin{equation}
A^{(t)} = \text{softmax} \left(X^{(l)} W_Q^{(l)} (X^{(l)} W_K^{(t)})^\top\right)
\end{equation}
\begin{equation}
X^{(t+1)} = W_O^{(l)}A^{(l)} X^{(l)} W_V^{(l)}
\end{equation}
\end{small}
where \( W_Q^{(l)}, W_K^{(l)}, W_V^{(l)}, W_O^{(l)}\in \mathbb{R}^{d \times d} \) are learnable projection matrices,  \( A^{(l)} \in \mathbb{R}^{n \times n} \) contains the attention weights and ${\sqrt{d_{QK}}}=1$ for simplicity.

\textbf{Assumptions.} To isolate the effect of position bias, we make simplifying assumptions:  (1) dominant first token attention,
(2)  position bias does not change over layers and (3) attention $A$ can be represented as a linear combination of contextual attention $A^{\mathrm{cont}}$ and positional attention \(A^{\mathrm{pos}}\).
% this assumption aligns with empirical observations that embeddings evolve slowly across layers~\citep{liu2023deja}.
    For all layers \( l \geq 1 \), the positional attention weights  $A_{i,j}^{pos_{(l)}}$ are sharply concentrated on the first token:
    
    $$
    A_{i,j}^{pos_{(l)}} \approx 
    \begin{cases}
    1 & \text{if } j = 1,\\  
    0 & \text{otherwise}.
    \end{cases}
    $$    
    Or in vector form $A^{pos_{(l)}} \approx \mathbf{1}_i^\top \mathbf{e}_1$. Linear combination of attention: $A =\lambda_1\,{A}^{\text{cont}} +\lambda_2\,{A}^{\text{pos}}, \qquad \lambda_1+\lambda_2 = 1; \lambda_1, \lambda_2 \in [0;1]$ where $\lambda$ are normalizing weights for each attention type.
    
    % This occurs when $W_Q^{(l)}$ and $WK^{(l)}$ align queries and keys such that $Q^{(l)} \cdot K_j^{(l)}$ is maximized for $j=1$. 
    % The value and output projection matrices approximate identity transformations:  $W_V^{(l)} \approx I, \quad W_O^{(l)} \approx I.$
   
    % 

\textbf{Token Homogenization.} Under these assumptions, the hidden state of token \( i \) at layer \( l \) becomes:
% \begin{small}
% \begin{equation}

% \end{equation}
% \end{small}

\begin{small}
\begin{equation}
\begin{aligned}
x_i^{(l)} 
&\approx W_O^{(l)} \cdot \left[ \sum_{j=1}^n \left( \lambda_1 A_{i,j}^{\text{pos}(l)} + \lambda_2 A_{i,j}^{\text{cont}(l)} \right) W_V^{(l)} x_j^{(l-1)} \right] \\
&= \lambda_2 W_O^{(l)} W_V^{(l)} \cdot \mathbf{1}_i^\top \mathbf{e}_1 x_i^{(l-1)} + \lambda_1 x_i^{\text{con}(l)} \\
&= \lambda_2 P^{(l)} x_i^{(l-1)} + \lambda_1 x_i^{\text{con}(l)},\forall i.
\end{aligned}
\end{equation}
\end{small}

Where $P^{(l)} = W_O^{(l)} W_V^{(l)}$. If $\lambda_2 > \lambda_1$ recursively applying this across $L$ layers yields:
\begin{equation}
x_i^{(L)} \approx x_1^{(0)} \quad \forall i,
\end{equation}

implying all tokens collapse to a copy of the initial first token embedding $x_1^{(0)}$ (up to projection transformations).
Although token collapse would not happen in a real scenario, for example, due to residual connections, tokens may become more similar to tokens under position bias.

\label{sec:entropy}
\textbf{Entropy Dynamics.} Let \( H_{A}^{(l)} \) denote the general attention entropy at layer \( l \). As tokens homogenize (\( x_i^{(l)} \approx x_1^{(0)} \)), queries and keys become indistinguishable, causing contextual attention weights to approach uniformity: \( A_{i,j}^{\text{con}_{(l)}} \approx \frac{1}{n} \quad \forall i,j \). This results in the maximization of entropy: \( H_{A}^{(l)} \to \log n \).

\textbf{Aligned positional and contextual attention.} This section establishes a theoretical connection between entropy, attention mechanisms, and position bias under several modeling assumptions. We demonstrate that alignment between contextual attention \(A^{\text{con}}\) and positional attention \(A^{\text{pos}}\) increases the likelihood of homogenization. This leads to a counterintuitive outcome: when the relevant context coincides with the model’s inherent positional bias, the model allocates greater attention to the corresponding tokens, ultimately elevating entropy. Further empirical support for these findings is provided in Section~\ref{sec:entropy}, where we observe that minimal entropy does not consistently coincide with alignment between contextual relevance and positional bias.

\textbf{Predictive and Attention Entropy.} Attention mechanisms are designed to prioritize relevant tokens within a sequence. When attention is uniformly distributed (indicating high entropy), the model is unable to effectively leverage contextual cues. This lack of focused attention results in diminished predictive signals, thereby increasing uncertainty in the model's output predictions (manifested as higher predictive entropy).\footnote{For example, in the sentence "The capital of France is ***," the token "France" is critical for accurate prediction. If attention is uniform, the model may assign equal weight to less relevant tokens (e.g., "The" or "of"), failing to emphasize "France." This ambiguity elevates uncertainty in predicting the subsequent token, such as "Paris."}

\textbf{Multilingual Caveat.} In multilingual large language models (LLMs), position bias may vary across different layers; for instance, early layers often prioritize English tokens, whereas later layers tend to align more closely with the language of the input prompt~\citep{zhong2024beyond, schut2025multilingual}. This shift may introduce language-specific positional preferences, thereby challenging the assumption of static position bias. Nevertheless, if homogenization occurs, the overall entropy dynamics remain consistent.

\section{Word Order Analysis}
\label{app:wordorder}
In this section, we perform additional investigation if there is  the relationship between the position of relevant context, model behavior, and the dominant word order of a language. Specifically, we examine whether position bias amplifies or diminishes the influence of a model's dominant language word order. Our analysis focuses on the \textit{No Scores} configuration with five contexts, using Hindi and German as representative languages due to their non-SVO dominant word orders, as documented in The World Atlas of Language Structures. Sentences are parsed using Stanza~\citep{qi2020stanza}, following a methodology similar to that of~\citep{choi-etal-2021-investigating}. For each verb, we identify its dependents; if a verb has both a subject (indicated by a "subj" relation, or such a relation for the nearest preceding verb connected via a "conj" dependency) and an object (indicated by an "obj" relation), we record the word order of these triplets using the abbreviations "S", "V", and "O". The distribution of "SVO" and "SOV" patterns is then analyzed relative to all extracted triplets.

\textbf{We find no evidence to suggest that position bias influences models to favor specific word orders.} For instance, in Hindi a predominantly subject–object–verb (SOV) language—one might expect subject–verb–object (SVO) rates to be lowest when relevant context appears last for Qwen and first for Llama, however, this pattern is not observed (Table~\ref{tab:word_order_positions}). In German, where dominant word order varies by clause type (SVO in main clauses and SOV in subordinate clauses; ~\citep{wals-81}), we examine the prevalence of SVO (quantified as the SVO–SOV difference) in Table~\ref{tab:svo_sov_complexity}. Here, also, no association with position bias is detected, as the observed prevalence appears to arise naturally from the proportion of complex sentences.

\definecolor{lightgreen}{RGB}{212,237,218}
\definecolor{lightred}{RGB}{237,212,218}
\begin{table}[ht!]
%\vspace{1mm}
\renewcommand{\arraystretch}{1.25}
\centering
\begingroup
  \small
  \setlength{\tabcolsep}{4pt}
\begin{tabularx}{\columnwidth}{ccZZZ}
\hhline{|-|-|-|-|-|}
\textbf{Model} & \textbf{Word order} & \textbf{TOP}  &  \textbf{MID} & \textbf{BOT}  \\
\hhline{|-|-|-|-|-|}
\multirow{2}{*}{Llama3.1 8B} 
  & SVO & \cellcolor{lightgreen} 5.39 & 5.12 & \textbf{4.75} \\
%\hhline{|~|-|-|-|-|}
  & SOV & \cellcolor{lightgreen} \textbf{88.52}  & 88.49 & 88.43 \\
\hhline{|-|-|-|-|-|}
\multirow{2}{*}{Qwen2.5 7B}
  & SVO & \textbf{2.34} & 4.15 & \cellcolor{lightgreen} 2.96 \\
%\hhline{|~|-|-|-|-|}
  & SOV & \textbf{93.98} & 91 & \cellcolor{lightgreen} 91.78  \\
\hhline{|-|-|-|-|-|}
\end{tabularx}
\endgroup
\caption{Percentage of word orders for Hindi, cells where context placements align with position bias are highlighted in green. Bolded results indicate stronger alignment with expected word order.}
\label{tab:word_order_positions}
\end{table}

% ==============================
% ===== Word Order German =====
% ==============================
\begin{table}[ht!]
%\vspace{1mm}
\centering
\begingroup
  \small
  \setlength{\tabcolsep}{4pt}
\renewcommand{\arraystretch}{1.25}
\begin{tabularx}{\columnwidth}{ccZZ}
\hline
% \rowcolor[gray]{0.85}
\multirow{1}{*}{\textbf{Model}} & \multirow{1}{*}{\textbf{Position}} & \textbf{Difference SVO - SOV} & \textbf{\% of Complex Sentences} \\
\hline
\multirow{3}{*}{Llama3.1 8B} 
  & TOP    & 26.88 & 13.56 \\ %\cline{2-4}
  & MIDDLE & 32.13 & 12.13 \\ %\cline{2-4}
  & BOTTOM & 40.04 & 10.88 \\ %\cline{2-4}
\hline
\multirow{3}{*}{Qwen2.5 7B} 
  & TOP    & 34.17 & 8.59 \\ %\cline{2-4}
  & MIDDLE & 29.56 & 9.54 \\ %\cline{2-4}
  & BOTTOM & 29.78 & 8.96 \\ %\cline{2-4}
\hline
\end{tabularx}
\endgroup
\caption{Prevalence of SVO over SOV word order and the proportion of complex sentences in German}
\label{tab:svo_sov_complexity}
\end{table}
For model-level analysis, the following distinctions are observed: (1) Llama generates a higher proportion of subject–verb–object (SVO) sentences than Qwen across Hindi, German, and Russian (Table~\ref{tab:svo_percentages})—note that Russian is included despite its typological preference for SVO order due to its rich morphological system, which permits considerable word-order flexibility~\citep{ghosh-etal-2024-morphology}; (2) Llama produces syntactically more complex sentences than Qwen across all three contextual positions (Table~\ref{tab:mean_complex_percentages}); (3) both models exhibit a tendency to generate more complex sentences when relevant context is provided in the initial position (Table~\ref{tab:context_position_complexity}).

% ==============================
% ===== Word Order SVO =====
% ==============================
\begin{table}[ht!]
%\vspace{1mm}
\renewcommand{\arraystretch}{1.25}
\centering
\begingroup
  \small
  \setlength{\tabcolsep}{4pt}
\begin{tabularx}{\columnwidth}{cZZZ}
\hline
%\rowcolor[gray]{0.85}
\textbf{Model} & \textbf{Hindi} & \textbf{German} & \textbf{Russian} \\
\hline
Llama3.1 8B & {\cellcolor{lightgreen} 4.8 - 5.4\%} & {\cellcolor{lightgreen} 60.3 - 67.5\%} & {\cellcolor{lightgreen} 95.9 - 96.2\%} \\
%\hline
Qwen2.5 7B & 2.3 - 4.2\% & 57.9 - 61.4\% & 93.6 - 94.5\% \\
\hline
\end{tabularx}
\endgroup
\caption{Percentage of SVO structures in generated responses across all three positions of relevant context}
\label{tab:svo_percentages}
\end{table}

% ==============================
% ===== Complex Sentences =====
% ==============================

\begin{table}[ht!]
%\vspace{1mm}
\renewcommand{\arraystretch}{1.25}
\centering
\begingroup
  \small
  \setlength{\tabcolsep}{4pt}
\begin{tabularx}{\columnwidth}{cZZZ}
\hline
\textbf{Model} & \textbf{Hindi} & \textbf{German} & \textbf{Russian} \\
\hline
Llama3.1 8B & {\cellcolor{lightgreen} 9.94 \%}  & {\cellcolor{lightgreen} 12.19 \%} & {\cellcolor{lightgreen} 7.19\%} \\
%\hline
Qwen2.5 7B & 6.88\% & 9.03 \% & 7\% \\
\hline
\end{tabularx}
\endgroup
\caption{Mean percentage of complex sentences among all sentences containing at least one triplet of a subject, verb, and object}
\label{tab:mean_complex_percentages}
\end{table}

% ==============================
% ===== Complex Sentences vs Context Position =====
% ==============================

\begin{table}[ht!]
%\vspace{1mm}
\renewcommand{\arraystretch}{1.25}
\centering
\begingroup
  \small
  \setlength{\tabcolsep}{4pt}
%\resizebox{.5\textwidth}{!}{%
\begin{tabularx}{\columnwidth}{l|Z|Z|Z|Z|Z|Z}
\hline
% \rowcolor[gray]{0.9}
 \multirow{2}{*}{\textbf{Language}} & \multicolumn{3}{c|}{\textbf{Llama3.1 8B}} & \multicolumn{3}{c}{\textbf{Qwen2.5 7B}} \\
\hhline{|~|-|-|-|-|-|-|}
% \rowcolor[gray]{0.9}
 & \textbf{TOP} & \textbf{MID} & \textbf{BOT} & \textbf{TOP} & \textbf{MID} & \textbf{BOT} \\
\hline
Hindi & $\checkmark$ &  &  & $\checkmark$ &  &  \\
\hline
German & $\checkmark$ &  &  &  & $\checkmark$ &  \\
\hline
Russian &  & $\checkmark$ &  & $\checkmark$ &  &  \\
\hline
\end{tabularx}
%}
\endgroup
\caption{The context position in which the proportion of complex sentences is highest, given a specific model and language}
\label{tab:context_position_complexity}
\end{table}

\section{Evaluation Details}
\label{app:lmjudge}

\subsection{LLM-as-a-Judge Verification}
\begin{figure*}[ht!]
\centering
\centerline{\includegraphics[width=12.0cm]{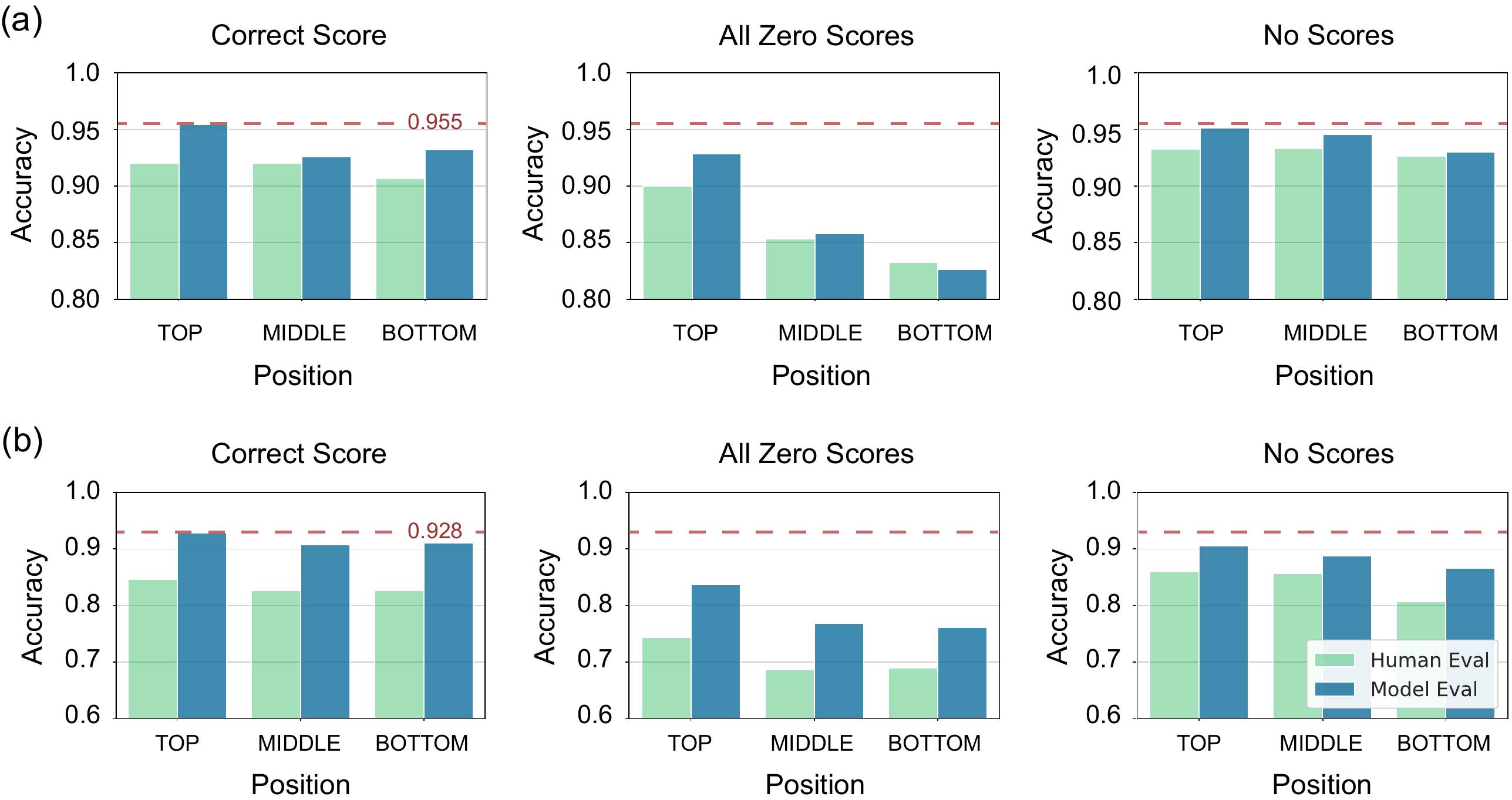}}
\caption{Human evaluation and LLM as a Judge for \textit{Correct Scores}(Aligned), \textit{All Zero}, and \textit{No Scores} strategies at three postions (TOP, MIDDLE, BOTTOM) for two languages: (a) English and (b) Russian. Bars in green represent human evaluations, while the blue bars represent the Llama model.}
\label{fig:human_vs_llm_judge}
\end{figure*}

Krippendorff’s alpha and Pearson correlation coefficients, calculated for each experimental setup can be seen in Tables \ref{tab:krippendorffs_alpha} and \ref{tab:pearson_corr} correspondingly. Comparison of human and Llama evaluation can be seen in Figure \ref{fig:human_vs_llm_judge}.

% =============================
% Human-LLM alignment (Krippendorffs alpha)
% =============================
\begin{table*}[ht!]
%\vspace{1mm}
\centering
\renewcommand{\arraystretch}{1.2}
\begin{tabularx}{\textwidth}{ccZZZZ}
\hhline{|-|-|-|-|-| |-|} 
\textbf{Language} & \textbf{Position} & \textbf{Aligned}  & \textbf{All Zero} & \textbf{No Scores} & \textbf{Mean} \\
\hhline{|-|-|-|-|-| |-|}  
\multirow{3}{*}{English} & TOP & {\cellcolor{lightgreen} 0.783} & 0.663 & 0.595 & \multirow{3}{*}{0.726} \\
%\hhline{|~|-|-|-|-| |~|}
& MIDDLE & 0.611 & 0.861 & 0.685 & \\
%\hhline{|~|-|-|-|-| |~|}
& BOTTOM & 0.704 & {\cellcolor{lightgreen} 0.916} & {\cellcolor{lightgreen} 0.718} & \\
\hhline{-----::-}
\multirow{3}{*}{Russian} & TOP & {\cellcolor{lightgreen} 0.814} & 0.825 & 0.674 & \multirow{3}{*}{0.783} \\
%\hhline{|~|-|-|-|-| |~|}
& MIDDLE & 0.742 & {\cellcolor{lightgreen} 0.865} & 0.695 & \\
%\hhline{|~|-|-|-|-| |~|}
& BOTTOM & 0.801 & 0.855 & {\cellcolor{lightgreen} 0.776} & \\
\hhline{|-|-|-|-|-| |-|} 
\end{tabularx}
\caption{Krippendorff's alpha coefficient, calculated for each experimental setup}
\label{tab:krippendorffs_alpha}
\end{table*}

% =============================
% Human-LLM alignment (Pearson)
% =============================
\begin{table*}[ht!]
%\vspace{1mm}
\centering
\renewcommand{\arraystretch}{1.2}
\begin{tabularx}{\textwidth}{ccZZZZ}
\hhline{|-|-|-|-|-| |-|} 
\textbf{Language} & \textbf{Position} & \textbf{Aligned}  & \textbf{All Zero} & \textbf{No Scores} & \textbf{Mean} \\
\hhline{|-|-|-|-|-| |-|}
\multirow{3}{*}{English} & TOP & 0.612 & 0.632 & 0.604 & \multirow{3}{*}{0.727} \\
%\hhline{|~|-|-|-|-| |~|}
& MIDDLE & {\cellcolor{lightgreen} 0.739} & 0.83 & {\cellcolor{lightgreen} 0.791} & \\
%\hhline{|~|-|-|-|-| |~|}
& BOTTOM & 0.738 & {\cellcolor{lightgreen} 0.908} & 0.685 & \\
\hhline{-----::-}
\multirow{3}{*}{Russian} & TOP & 0.488 & 0.704 & 0.518 & \multirow{3}{*}{0.705} \\
%\hhline{|~|-|-|-|-| |~|}
& MIDDLE & {\cellcolor{lightgreen} 0.709} & 0.87 & {\cellcolor{lightgreen} 0.769} & \\
%\hhline{|~|-|-|-|-| |~|}
& BOTTOM & 0.669 & {\cellcolor{lightgreen} 0.888} & 0.732 & \\
\hhline{|-|-|-|-|-| |-|} 
\end{tabularx}
\caption{Pearson correlation coefficient, calculated for each experimental setup}
\label{tab:pearson_corr}
\end{table*}

\subsection{Human and LLM instructions}
We prompt Mistral-Large to judge whether the LLM responses are correctly answering questions. For each dataset we create an evaluation prompt on the language of this dataset and add 4 shots as examples of judgments. The resulting prompt consists of \textit{system prompt} "You are an AI assistant who speaks English.", which we translate to other languages and \textit{user prompt}. %User prompts for each language could be found in Table \ref{tab:prompts_judge}.

For human annotation we consider only English and Russian languages, since our annotators speaks these languages. We use the same instructions as for LLM-as-a-Judge settings, omitting shots.

\subsection{Human Annotators Information}
Annotation was conducted by the authors of the work, so no additional recruitment or payment are required on this stage. All assessors held bachelor’s degree and had prior experience in the evaluation of LLM responses.

\section{Context Volume Analysis}
\label{app:number_of_prompts}
\begin{figure*}[th!]
\centering
\centerline{\includegraphics[width=13.0cm]{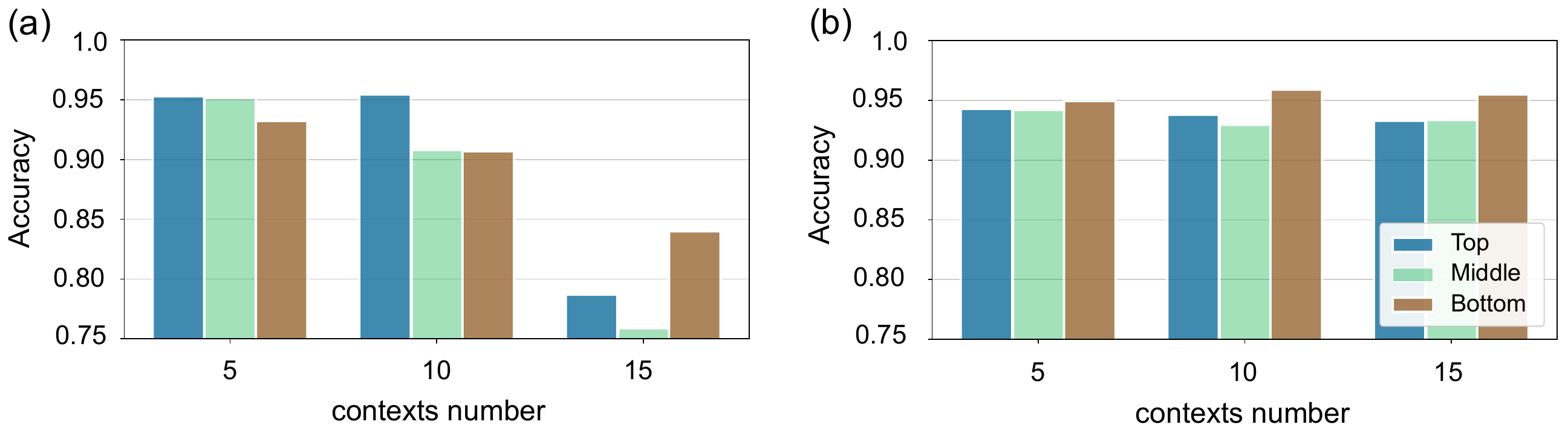}}
\caption{Accuracy dependence on the number of contexts, added to the user-prompt, and position of the relevant context in a list with \textit{Aligned} placement strategy: (a) Llama3.1-8B; (b) Qwen2.5-7B}
\label{fig:contexts_position}
\end{figure*}

From Figure \ref{fig:contexts_position} we can observe that our \textit{Aligned} strategy does not have an effect on position bias with increasing of information load. With context quantity $N=15$ for Llama3.1-8B we can see significant accuracy decrease, compared  to other quantities. From the other hand, for Qwen2.5-7B position bias does not correlate with passed number of contexts. This result can be explained by the fact that for Qwen2.5-7B training larger dataset with long contexts was used, compared to Llama3.1-8B. This feature increase for Qwen2.5-7B the size of its attention window and allowed to conditioning on a larger amount of input knowledge during response generation.

%\clearpage

\section{Non Aggregated results for main experiments}
\label{app:extres}

Our non aggregated results for LLM as an accuracy and entropy are presented in Tables~\ref{tab:llmjudge_full1},\ref{tab:llmjudge_full2} and \ref{tab:entropy_full1}, \ref{tab:entropy_full2} correspondingly.

\begin{table*}[ht!]
\centering
%\vspace{1mm}
\resizebox{\textwidth}{!}{%
\renewcommand{\arraystretch}{1.3}
\begin{tabular}{ccccc|ccc|ccc}
\hhline{-----|---|---}
 &  & \multicolumn{3}{c|}{\textbf{DeepSeek 7B}} & \multicolumn{3}{c|}{\textbf{Gemma 7B}} & \multicolumn{3}{c}{\textbf{Mistral 7B}} \\ 
 %\hhline{|~|~||---||---||---|}
\multirow{-2}{*}{\textbf{Dataset}} & \multirow{-2}{*}{\textbf{Position}} & \multicolumn{1}{c}{\textbf{Aligned}} & \multicolumn{1}{c}{\textbf{All Zero}} & \textbf{No Scores} & \multicolumn{1}{c}{\textbf{Aligned}} & \multicolumn{1}{c}{\textbf{All Zero}} & \textbf{No Scores} & \multicolumn{1}{c}{\textbf{Aligned}} & \multicolumn{1}{c}{\textbf{All Zero}} & \textbf{No Scores} \\ 
\hhline{-----|---|---}
 & TOP & \multicolumn{1}{c}{\colorcell{0.907}} & \multicolumn{1}{c}{\colorcell{0.877}} & \colorcell{0.913} & \multicolumn{1}{c}{\colorcell{0.815}} & \multicolumn{1}{c}{\colorcell{0.683}} & \colorcell{0.831} & \multicolumn{1}{c}{\colorcell{0.970}} & \multicolumn{1}{c}{\colorcell{0.956}} & \colorcell{0.961} \\ 
 %\hhline{|~|-||---||---||---|} 
 & MIDDLE & \multicolumn{1}{c}{\colorcell{0.922}} & \multicolumn{1}{c}{\colorcell{0.887}} & \colorcell{0.921} & \multicolumn{1}{c}{\colorcell{0.849}} & \multicolumn{1}{c}{\colorcell{0.614}} & \colorcell{0.832} & \multicolumn{1}{c}{\colorcell{0.958}} & \multicolumn{1}{c}{\colorcell{0.929}} & \colorcell{0.957} \\
 %\hhline{|~|-||---||---||---|} 
\multirow{-3}{*}{Squadv2 (en)} & BOTTOM & \multicolumn{1}{c}{\colorcell{0.953}} & \multicolumn{1}{c}{\colorcell{0.929}} & \colorcell{0.950} & \multicolumn{1}{c}{\colorcell{0.782}} & \multicolumn{1}{c}{\colorcell{0.534}} & \colorcell{0.815} & \multicolumn{1}{c}{\colorcell{0.963}} & \multicolumn{1}{c}{\colorcell{0.939}} & \colorcell{0.958} \\ 
\hhline{-----|---|---}
 & TOP & \multicolumn{1}{c}{\colorcell{0.804}} & \multicolumn{1}{c}{\colorcell{0.791}} & \colorcell{0.841} & \multicolumn{1}{c}{\colorcell{0.775}} & \multicolumn{1}{c}{\colorcell{0.655}} & \colorcell{0.797} & \multicolumn{1}{c}{\colorcell{0.939}} & \multicolumn{1}{c}{\colorcell{0.926}} & \colorcell{0.936} \\ 
 %\hhline{|~|-||---||---||---|} 
 & MIDDLE & \multicolumn{1}{c}{\colorcell{0.834}} & \multicolumn{1}{c}{\colorcell{0.829}} & \colorcell{0.843} & \multicolumn{1}{c}{\colorcell{0.777}} & \multicolumn{1}{c}{\colorcell{0.701}} & \colorcell{0.819} & \multicolumn{1}{c}{\colorcell{0.920}} & \multicolumn{1}{c}{\colorcell{0.911}} & \colorcell{0.924} \\ 
%\hhline{|~|-||---||---||---|} 
\multirow{-3}{*}{MTS-SQuAD (ru)} & BOTTOM & \multicolumn{1}{c}{\colorcell{0.884}} & \multicolumn{1}{c}{\colorcell{0.875}} & \colorcell{0.881} & \multicolumn{1}{c}{\colorcell{0.782}} & \multicolumn{1}{c}{\colorcell{0.707}} & \colorcell{0.824} & \multicolumn{1}{c}{\colorcell{0.932}} & \multicolumn{1}{c}{\colorcell{0.928}} & \colorcell{0.930} \\
\hhline{-----|---|---}
 & TOP & \multicolumn{1}{c}{\colorcell{0.737}} & \multicolumn{1}{c}{\colorcell{0.732}} & \colorcell{0.720} & \multicolumn{1}{c}{\colorcell{0.477}} & \multicolumn{1}{c}{\colorcell{0.324}} & \colorcell{0.524} & \multicolumn{1}{c}{\colorcell{0.829}} & \multicolumn{1}{c}{\colorcell{0.799}} & \colorcell{0.813} \\ 
 %\hhline{|~|-||---||---||---|}
 & MIDDLE & \multicolumn{1}{c}{\colorcell{0.735}} & \multicolumn{1}{c}{\colorcell{0.708}} & \colorcell{0.724} & \multicolumn{1}{c}{\colorcell{0.508}} & \multicolumn{1}{c}{\colorcell{0.289}} & \colorcell{0.486} & \multicolumn{1}{c}{\colorcell{0.815}} & \multicolumn{1}{c}{\colorcell{0.750}} & \colorcell{0.801} \\ 
 %\hhline{|~|-||---||---||---|}
\multirow{-3}{*}{MLQA (de)} & BOTTOM & \multicolumn{1}{c}{\colorcell{0.799}} & \multicolumn{1}{c}{\colorcell{0.772}} & \colorcell{0.777} & \multicolumn{1}{c}{\colorcell{0.432}} & \multicolumn{1}{c}{\colorcell{0.244}} & \colorcell{0.441} & \multicolumn{1}{c}{\colorcell{0.823}} & \multicolumn{1}{c}{\colorcell{0.787}} & \colorcell{0.819} \\
\hhline{-----|---|---}
 & TOP & \multicolumn{3}{c|}{\cellcolor{lightred}} & \multicolumn{1}{c}{\colorcell{0.431}} & \multicolumn{1}{c}{\colorcell{0.372}} & \colorcell{0.229} & \multicolumn{1}{c}{\colorcell{0.520}} & \multicolumn{1}{c}{\colorcell{0.439}} & \colorcell{0.601} \\ 
 %\hhline{|~|-||~~~||---||---|} %\cline{2-2} \cline{6-11}  
 & MIDDLE & \multicolumn{3}{c|}{\cellcolor{lightred}} & \multicolumn{1}{c}{\colorcell{0.402}} & \multicolumn{1}{c}{\colorcell{0.334}} & \colorcell{0.237} & \multicolumn{1}{c}{\colorcell{0.505}} & \multicolumn{1}{c}{\colorcell{0.425}} & \colorcell{0.558} \\ 
 %\hhline{|~|-||~~~||---||---|} %\cline{2-2} \cline{6-11} 
\multirow{-3}{*}{MLQA (hi)} & BOTTOM & \multicolumn{3}{c|}{\multirow{-3}{*}{\cellcolor{lightred}-}} & \multicolumn{1}{c}{\colorcell{0.393}} & \multicolumn{1}{c}{\colorcell{0.349}} & \colorcell{0.228} & \multicolumn{1}{c}{\colorcell{0.546}} & \multicolumn{1}{c}{\colorcell{0.471}} & \colorcell{0.585} \\ 
\hhline{-----|---|---}
 & TOP & \multicolumn{1}{c}{\colorcell{0.284}} & \multicolumn{1}{c}{\colorcell{0.277}} & \colorcell{0.309} & \multicolumn{1}{c}{\colorcell{0.039}} & \multicolumn{1}{c}{\colorcell{0.021}} & \colorcell{0.484} & \multicolumn{1}{c}{\colorcell{0.520}} & \multicolumn{1}{c}{\colorcell{0.428}} & \colorcell{0.584} \\ 
 %\hhline{|~|-||---||---||---|}
 & MIDDLE & \multicolumn{1}{c}{\colorcell{0.317}} & \multicolumn{1}{c}{\colorcell{0.314}} & \colorcell{0.330} & \multicolumn{1}{c}{\colorcell{0.039}} & \multicolumn{1}{c}{\colorcell{0.020}} & \colorcell{0.483} & \multicolumn{1}{c}{\colorcell{0.517}} & \multicolumn{1}{c}{\colorcell{0.406}} & \colorcell{0.566} \\ 
 %\hhline{|~|-||---||---||---|}
\multirow{-3}{*}{MLQA (vi)} & BOTTOM & \multicolumn{1}{c}{\colorcell{0.407}} & \multicolumn{1}{c}{\colorcell{0.397}} & \colorcell{0.415} & \multicolumn{1}{c}{\colorcell{0.035}} & \multicolumn{1}{c}{\colorcell{0.014}} & \colorcell{0.471} & \multicolumn{1}{c}{\colorcell{0.545}} & \multicolumn{1}{c}{\colorcell{0.457}} & \colorcell{0.588} \\ \hline
\end{tabular}%
}
\caption{Accuracy for three models: DeepSeek 7B, Gemma 7B and Mistral 7B, DeepSeek consistently failed in Hindi}
\label{tab:llmjudge_full1}
\end{table*}

\begin{table*}[ht!]
\centering
%\vspace{1mm}
\resizebox{0.8\textwidth}{!}{%
\renewcommand{\arraystretch}{1.3}
\begin{tabular}{ccccc|ccc}
\hline
\multirow{2}{*}{\textbf{Dataset}} & \multirow{2}{*}{\textbf{Position}} & \multicolumn{3}{c|}{\textbf{Llama3.1 8B}} & \multicolumn{3}{c}{\textbf{Qwen2.5 7B}} \\ 
%\hhline{|~|~||---||---|}
 &  & \multicolumn{1}{c}{\textbf{Aligned}} & \multicolumn{1}{c}{\textbf{All Zero}} & \textbf{No Scores} & \multicolumn{1}{c}{\textbf{Aligned}} & \multicolumn{1}{c}{\textbf{All Zero}} & \textbf{No Scores} \\ 
 \hhline{-----|---}
\multirow{3}{*}{SQuAD2.0 (en)} & TOP & \multicolumn{1}{c}{\colorcell{0.955}} & \multicolumn{1}{c}{\colorcell{0.929}} & \colorcell{0.951} & \multicolumn{1}{c}{\colorcell{0.943}} & \multicolumn{1}{c}{\colorcell{0.927}} & \colorcell{0.952} \\ 
%\hhline{|~|-||---||---|}
 & MIDDLE & \multicolumn{1}{c}{\colorcell{0.926}} & \multicolumn{1}{c}{\colorcell{0.858}} & \colorcell{0.946} & \multicolumn{1}{c}{\colorcell{0.941}} & \multicolumn{1}{c}{\colorcell{0.931}} & \colorcell{0.945} \\
 %\hhline{|~|-||---||---|}
 & BOTTOM & \multicolumn{1}{c}{\colorcell{0.932}} & \multicolumn{1}{c}{\colorcell{0.826}} & \colorcell{0.930} & \multicolumn{1}{c}{\colorcell{0.949}} & \multicolumn{1}{c}{\colorcell{0.948}} & \colorcell{0.956} \\ 
 \hhline{-----|---}
\multirow{3}{*}{MTS-SQuAD  (ru)} & TOP & \multicolumn{1}{c}{\colorcell{0.928}} & \multicolumn{1}{c}{\colorcell{0.837}} & \colorcell{0.906} & \multicolumn{1}{c}{\colorcell{0.938}} & \multicolumn{1}{c}{\colorcell{0.909}} & \colorcell{0.934} \\ 
%\hhline{|~|-||---||---|}
 & MIDDLE & \multicolumn{1}{c}{\colorcell{0.908}} & \multicolumn{1}{c}{\colorcell{0.768}} & \colorcell{0.888} & \multicolumn{1}{c}{\colorcell{0.927}} & \multicolumn{1}{c}{\colorcell{0.917}} & \colorcell{0.944} \\ 
% \hhline{|~|-||---||---|}
 & BOTTOM & \multicolumn{1}{c}{\colorcell{0.911}} & \multicolumn{1}{c}{\colorcell{0.761}} & \colorcell{0.866} & \multicolumn{1}{c}{\colorcell{0.937}} & \multicolumn{1}{c}{\colorcell{0.935}} & \colorcell{0.939} \\ 
 \hhline{-----|---}
\multirow{3}{*}{MLQA (de)} & TOP & \multicolumn{1}{c}{\colorcell{0.680}} & \multicolumn{1}{c}{\colorcell{0.485}} & \colorcell{0.719} & \multicolumn{1}{c}{\colorcell{0.648}} & \multicolumn{1}{c}{\colorcell{0.553}} & \colorcell{0.644} \\ 
%\hhline{|~|-||---||---|}
 & MIDDLE & \multicolumn{1}{c}{\colorcell{0.602}} & \multicolumn{1}{c}{\colorcell{0.312}} & \colorcell{0.668} & \multicolumn{1}{c}{\colorcell{0.629}} & \multicolumn{1}{c}{\colorcell{0.566}} & \colorcell{0.628} \\ 
% \hhline{|~|-||---||---|}
 & BOTTOM & \multicolumn{1}{c}{\colorcell{0.613}} & \multicolumn{1}{c}{\colorcell{0.307}} & \colorcell{0.646} & \multicolumn{1}{c}{\colorcell{0.655}} & \multicolumn{1}{c}{\colorcell{0.602}} & \colorcell{0.649} \\ 
 \hhline{-----|---}
\multirow{3}{*}{MLQA (hi)} & TOP & \multicolumn{1}{c}{\colorcell{0.532}} & \multicolumn{1}{c}{\colorcell{0.309}} & \colorcell{0.729} & \multicolumn{1}{c}{\colorcell{0.591}} & \multicolumn{1}{c}{\colorcell{0.540}} & \colorcell{0.619} \\
%\hhline{|~|-||---||---|}
 & MIDDLE & \multicolumn{1}{c}{\colorcell{0.448}} & \multicolumn{1}{c}{\colorcell{0.112}} & \colorcell{0.676} & \multicolumn{1}{c}{\colorcell{0.579}} & \multicolumn{1}{c}{\colorcell{0.543}} & \colorcell{0.609} \\ 
% \hhline{|~|-||---||---|}
 & BOTTOM & \multicolumn{1}{c}{\colorcell{0.494}} & \multicolumn{1}{c}{\colorcell{0.038}} & \colorcell{0.644} & \multicolumn{1}{c}{\colorcell{0.642}} & \multicolumn{1}{c}{\colorcell{0.592}} & \colorcell{0.643} \\ 
 \hhline{-----|---}
\multirow{3}{*}{MLQA (vi)} & TOP & \multicolumn{1}{c}{\colorcell{0.764}} & \multicolumn{1}{c}{\colorcell{0.679}} & \colorcell{0.737} & \multicolumn{1}{c}{\colorcell{0.722}} & \multicolumn{1}{c}{\colorcell{0.644}} & \colorcell{0.713} \\
%\hhline{|~|-||---||---|}
 & MIDDLE & \multicolumn{1}{c}{\colorcell{0.714}} & \multicolumn{1}{c}{\colorcell{0.536}} & \colorcell{0.707} & \multicolumn{1}{c}{\colorcell{0.676}} & \multicolumn{1}{c}{\colorcell{0.619}} & \colorcell{0.707} \\ 
% \hhline{|~|-||---||---|}
 & BOTTOM & \multicolumn{1}{c}{\colorcell{0.726}} & \multicolumn{1}{c}{\colorcell{0.473}} & \colorcell{0.656} & \multicolumn{1}{c}{\colorcell{0.718}} & \multicolumn{1}{c}{\colorcell{0.647}} & \colorcell{0.700} \\ \hline
\end{tabular}%
}
\caption{Accuracy for two models: Llama3.1 8B and Qwen2.5 7B}
\label{tab:llmjudge_full2}
\end{table*}

\begin{table*}[ht]
\centering
%\vspace{1mm}
\resizebox{\textwidth}{!}{%
\renewcommand{\arraystretch}{1.3}
\begin{tabular}{ccccc|ccc|ccc}
\hline
 &  & \multicolumn{3}{c|}{\textbf{DeepSeek 7B Chat}} & \multicolumn{3}{c|}{\textbf{Gemma 7B}} & \multicolumn{3}{c}{\textbf{Mistral 7B}} \\ 
%\hhline{|~|~||---||---||---|}
\multirow{-2}{*}{\textbf{Dataset}} & \multirow{-2}{*}{\textbf{Position}} & \multicolumn{1}{c}{\textbf{Aligned}} & \multicolumn{1}{c}{\textbf{All Zero}} & \textbf{No Scores} & \multicolumn{1}{c}{\textbf{Aligned}} & \multicolumn{1}{c}{\textbf{All Zero}} & \textbf{No Scores} & \multicolumn{1}{c}{\textbf{Aligned}} & \multicolumn{1}{c}{\textbf{All Zero}} & \textbf{No Scores} \\ 
\hhline{-----|---|---}
 & TOP & \multicolumn{1}{c}{\colorcellentropy{0.140}} & \multicolumn{1}{c}{\colorcellentropy{0.148}} & \colorcellentropy{0.121} & \multicolumn{1}{c}{\colorcellentropy{0.131}} & \multicolumn{1}{c}{\colorcellentropy{0.160}} & \colorcellentropy{0.116} & \multicolumn{1}{c}{\colorcellentropy{0.122}} & \multicolumn{1}{c}{\colorcellentropy{0.124}} & \colorcellentropy{0.126} \\ 
%\hhline{|~|-||---||---||---|}
 & MIDDLE & \multicolumn{1}{c}{\colorcellentropy{0.132}} & \multicolumn{1}{c}{\colorcellentropy{0.137}} & \colorcellentropy{0.116} & \multicolumn{1}{c}{\colorcellentropy{0.141}} & \multicolumn{1}{c}{\colorcellentropy{0.166}} & \colorcellentropy{0.110} & \multicolumn{1}{c}{\colorcellentropy{0.125}} & \multicolumn{1}{c}{\colorcellentropy{0.130}} & \colorcellentropy{0.131} \\ 
%\hhline{|~|-||---||---||---|}
\multirow{-3}{*}{SQuAD2.0 (en)} & BOTTOM & \multicolumn{1}{c}{\colorcellentropy{0.128}} & \multicolumn{1}{c}{\colorcellentropy{0.129}} & \colorcellentropy{0.114} & \multicolumn{1}{c}{\colorcellentropy{0.146}} & \multicolumn{1}{c}{\colorcellentropy{0.175}} & \colorcellentropy{0.113} & \multicolumn{1}{c}{\colorcellentropy{0.117}} & \multicolumn{1}{c}{\colorcellentropy{0.122}} & \colorcellentropy{0.125} \\ 
\hhline{-----|---|---}
 & TOP & \multicolumn{1}{c}{\colorcellentropy{0.225}} & \multicolumn{1}{c}{\colorcellentropy{0.222}} & \colorcellentropy{0.240} & \multicolumn{1}{c}{\colorcellentropy{0.183}} & \multicolumn{1}{c}{\colorcellentropy{0.229}} & \colorcellentropy{0.155} & \multicolumn{1}{c}{\colorcellentropy{0.185}} & \multicolumn{1}{c}{\colorcellentropy{0.191}} & \colorcellentropy{0.204} \\
% \hhline{|~|-||---||---||---|}
 & MIDDLE & \multicolumn{1}{c}{\colorcellentropy{0.216}} & \multicolumn{1}{c}{\colorcellentropy{0.222}} & \colorcellentropy{0.232} & \multicolumn{1}{c}{\colorcellentropy{0.164}} & \multicolumn{1}{c}{\colorcellentropy{0.204}} & \colorcellentropy{0.149} & \multicolumn{1}{c}{\colorcellentropy{0.193}} & \multicolumn{1}{c}{\colorcellentropy{0.200}} & \colorcellentropy{0.210} \\
% \hhline{|~|-||---||---||---|}
\multirow{-3}{*}{MTS-SQuAD  (ru)} & BOTTOM & \multicolumn{1}{c}{\colorcellentropy{0.208}} & \multicolumn{1}{c}{\colorcellentropy{0.212}} & \colorcellentropy{0.231} & \multicolumn{1}{c}{\colorcellentropy{0.167}} & \multicolumn{1}{c}{\colorcellentropy{0.199}} & \colorcellentropy{0.145} & \multicolumn{1}{c}{\colorcellentropy{0.190}} & \multicolumn{1}{c}{\colorcellentropy{0.197}} & \colorcellentropy{0.207} \\ 
\hhline{-----|---|---}
 & TOP & \multicolumn{1}{c}{\colorcellentropy{0.237}} & \multicolumn{1}{c}{\colorcellentropy{0.241}} & \colorcellentropy{0.240} & \multicolumn{1}{c}{\colorcellentropy{0.234}} & \multicolumn{1}{c}{\colorcellentropy{0.228}} & \colorcellentropy{0.219} & \multicolumn{1}{c}{\colorcellentropy{0.276}} & \multicolumn{1}{c}{\colorcellentropy{0.288}} & \colorcellentropy{0.292} \\ 
%\hhline{|~|-||---||---||---|}
 & MIDDLE & \multicolumn{1}{c}{\colorcellentropy{0.238}} & \multicolumn{1}{c}{\colorcellentropy{0.249}} & \colorcellentropy{0.238} & \multicolumn{1}{c}{\colorcellentropy{0.218}} & \multicolumn{1}{c}{\colorcellentropy{0.209}} & \colorcellentropy{0.225} & \multicolumn{1}{c}{\colorcellentropy{0.286}} & \multicolumn{1}{c}{\colorcellentropy{0.321}} & \colorcellentropy{0.295} \\ 
%\hhline{|~|-||---||---||---|}
\multirow{-3}{*}{MLQA (de)} & BOTTOM & \multicolumn{1}{c}{\colorcellentropy{0.222}} & \multicolumn{1}{c}{\colorcellentropy{0.233}} & \colorcellentropy{0.228} & \multicolumn{1}{c}{\colorcellentropy{0.215}} & \multicolumn{1}{c}{\colorcellentropy{0.197}} & \colorcellentropy{0.235} & \multicolumn{1}{c}{\colorcellentropy{0.274}} & \multicolumn{1}{c}{\colorcellentropy{0.300}} & \colorcellentropy{0.287} \\ 
\hhline{-----|---|---}
 & TOP & \multicolumn{3}{c|}{\cellcolor{lightred}} & \multicolumn{1}{c}{\colorcellentropy{0.279}} & \multicolumn{1}{c}{\colorcellentropy{0.325}} & \colorcellentropy{0.366} & \multicolumn{1}{c}{\colorcellentropy{0.213}} & \multicolumn{1}{c}{\colorcellentropy{0.202}} & \colorcellentropy{0.238} \\ 
%\hhline{|~|-||~~~||---||---|}
 & MIDDLE & \multicolumn{3}{c|}{\cellcolor{lightred}} & \multicolumn{1}{c}{\colorcellentropy{0.302}} & \multicolumn{1}{c}{\colorcellentropy{0.349}} & \colorcellentropy{0.382} & \multicolumn{1}{c}{\colorcellentropy{0.221}} & \multicolumn{1}{c}{\colorcellentropy{0.204}} & \colorcellentropy{0.236} \\ 
%\hhline{|~|-||~~~||---||---|}
\multirow{-3}{*}{MLQA (hi)} & BOTTOM & \multicolumn{3}{c|}{\multirow{-3}{*}{\cellcolor{lightred}-}} & \multicolumn{1}{c}{\colorcellentropy{0.312}} & \multicolumn{1}{c}{\colorcellentropy{0.343}} & \colorcellentropy{0.391} & \multicolumn{1}{c}{\colorcellentropy{0.227}} & \multicolumn{1}{c}{\colorcellentropy{0.214}} & \colorcellentropy{0.237} \\ 
\hhline{-----|---|---}
 & TOP & \multicolumn{1}{c}{\colorcellentropy{0.427}} & \multicolumn{1}{c}{\colorcellentropy{0.423}} & \colorcellentropy{0.332} & \multicolumn{1}{c}{\colorcellentropy{0.120}} & \multicolumn{1}{c}{\colorcellentropy{0.112}} & \colorcellentropy{0.115} & \multicolumn{1}{c}{\colorcellentropy{0.175}} & \multicolumn{1}{c}{\colorcellentropy{0.166}} & \colorcellentropy{0.225} \\
%\hhline{|~|-||---||---||---|}
 & MIDDLE & \multicolumn{1}{c}{\colorcellentropy{0.412}} & \multicolumn{1}{c}{\colorcellentropy{0.409}} & \colorcellentropy{0.322} & \multicolumn{1}{c}{\colorcellentropy{0.110}} & \multicolumn{1}{c}{\colorcellentropy{0.106}} & \colorcellentropy{0.113} & \multicolumn{1}{c}{\colorcellentropy{0.185}} & \multicolumn{1}{c}{\colorcellentropy{0.171}} & \colorcellentropy{0.232} \\ 
%\hhline{|~|-||---||---||---|}
\multirow{-3}{*}{MLQA (vi)} & BOTTOM & \multicolumn{1}{c}{\colorcellentropy{0.374}} & \multicolumn{1}{c}{\colorcellentropy{0.379}} & \colorcellentropy{0.277} & \multicolumn{1}{c}{\colorcellentropy{0.104}} & \multicolumn{1}{c}{\colorcellentropy{0.104}} & \colorcellentropy{0.114} & \multicolumn{1}{c}{\colorcellentropy{0.185}} & \multicolumn{1}{c}{\colorcellentropy{0.171}} & \colorcellentropy{0.218} \\ \hline
\end{tabular}%
}
\caption{Predictive entropy values for three models: DeepSeek 7B Chat, Gemma 7B and Mistral 7B,  Deep Seek consistently failed in Hindi}
\label{tab:entropy_full1}
\end{table*}

\begin{table*}[ht]
\centering
%\vspace{1mm}
\resizebox{0.8\textwidth}{!}{%
\renewcommand{\arraystretch}{1.3}
\begin{tabular}{ccccc|ccc}
\hline
\multirow{2}{*}{\textbf{Dataset}} & \multirow{2}{*}{\textbf{Position}} & \multicolumn{3}{c|}{\textbf{Llama3.1 8B}} & \multicolumn{3}{c}{\textbf{Qwen2.5 7B}} \\ 
%\hhline{|~|~||---||---|}
 &  & \multicolumn{1}{c}{\textbf{Aligned}} & \multicolumn{1}{c}{\textbf{All Zero}} & \textbf{No Scores} & \multicolumn{1}{c}{\textbf{Aligned}} & \multicolumn{1}{c}{\textbf{All Zero}} & \textbf{No Scores} \\ 
\hhline{-----|---}
\multirow{3}{*}{SQuAD2.0 (en)} & TOP & \multicolumn{1}{c}{\colorcellentropy{0.232}} & \multicolumn{1}{c}{\colorcellentropy{0.248}} & \colorcellentropy{0.194} & \multicolumn{1}{c}{\colorcellentropy{0.092}} & \multicolumn{1}{c}{\colorcellentropy{0.093}} & \colorcellentropy{0.100} \\ 
%\hhline{|~|-||---||---|}
 & MIDDLE & \multicolumn{1}{c}{\colorcellentropy{0.237}} & \multicolumn{1}{c}{\colorcellentropy{0.239}} & \colorcellentropy{0.201} & \multicolumn{1}{c}{\colorcellentropy{0.094}} & \multicolumn{1}{c}{\colorcellentropy{0.102}} & \colorcellentropy{0.100} \\ 
%\hhline{|~|-||---||---|}
 & BOTTOM & \multicolumn{1}{c}{\colorcellentropy{0.241}} & \multicolumn{1}{c}{\colorcellentropy{0.240}} & \colorcellentropy{0.206} & \multicolumn{1}{c}{\colorcellentropy{0.093}} & \multicolumn{1}{c}{\colorcellentropy{0.099}} & \colorcellentropy{0.097} \\ 
\hhline{-----|---}
\multirow{3}{*}{MTS-SQuAD  (ru)} & TOP & \multicolumn{1}{c}{\colorcellentropy{0.203}} & \multicolumn{1}{c}{\colorcellentropy{0.224}} & \colorcellentropy{0.169} & \multicolumn{1}{c}{\colorcellentropy{0.105}} & \multicolumn{1}{c}{\colorcellentropy{0.111}} & \colorcellentropy{0.129} \\ 
%\hhline{|~|-||---||---|}
 & MIDDLE & \multicolumn{1}{c}{\colorcellentropy{0.217}} & \multicolumn{1}{c}{\colorcellentropy{0.219}} & \colorcellentropy{0.182} & \multicolumn{1}{c}{\colorcellentropy{0.112}} & \multicolumn{1}{c}{\colorcellentropy{0.118}} & \colorcellentropy{0.126} \\
%\hhline{|~|-||---||---|}
 & BOTTOM & \multicolumn{1}{c}{\colorcellentropy{0.214}} & \multicolumn{1}{c}{\colorcellentropy{0.222}} & \colorcellentropy{0.191} & \multicolumn{1}{c}{\colorcellentropy{0.112}} & \multicolumn{1}{c}{\colorcellentropy{0.109}} & \colorcellentropy{0.126} \\ 
\hhline{-----|---}
\multirow{3}{*}{MLQA (de)} & TOP & \multicolumn{1}{c}{\colorcellentropy{0.231}} & \multicolumn{1}{c}{\colorcellentropy{0.204}} & \colorcellentropy{0.220} & \multicolumn{1}{c}{\colorcellentropy{0.117}} & \multicolumn{1}{c}{\colorcellentropy{0.106}} & \colorcellentropy{0.115} \\ 
%\hhline{|~|-||---||---|}
 & MIDDLE & \multicolumn{1}{c}{\colorcellentropy{0.220}} & \multicolumn{1}{c}{\colorcellentropy{0.158}} & \colorcellentropy{0.225} & \multicolumn{1}{c}{\colorcellentropy{0.126}} & \multicolumn{1}{c}{\colorcellentropy{0.118}} & \colorcellentropy{0.130} \\ 
 %\hhline{|~|-||---||---|}
 & BOTTOM & \multicolumn{1}{c}{\colorcellentropy{0.225}} & \multicolumn{1}{c}{\colorcellentropy{0.152}} & \colorcellentropy{0.220} & \multicolumn{1}{c}{\colorcellentropy{0.128}} & \multicolumn{1}{c}{\colorcellentropy{0.118}} & \colorcellentropy{0.125} \\ 
\hhline{-----|---}
\multirow{3}{*}{MLQA (hi)} & TOP & \multicolumn{1}{c}{\colorcellentropy{0.303}} & \multicolumn{1}{c}{\colorcellentropy{0.320}} & \colorcellentropy{0.237} & \multicolumn{1}{c}{\colorcellentropy{0.078}} & \multicolumn{1}{c}{\colorcellentropy{0.075}} & \colorcellentropy{0.076} \\ 
%\hhline{|~|-||---||---|}
 & MIDDLE & \multicolumn{1}{c}{\colorcellentropy{0.294}} & \multicolumn{1}{c}{\colorcellentropy{0.290}} & \colorcellentropy{0.257} & \multicolumn{1}{c}{\colorcellentropy{0.082}} & \multicolumn{1}{c}{\colorcellentropy{0.077}} & \colorcellentropy{0.082} \\ 
%\hhline{|~|-||---||---|}
 & BOTTOM & \multicolumn{1}{c}{\colorcellentropy{0.289}} & \multicolumn{1}{c}{\colorcellentropy{0.277}} & \colorcellentropy{0.275} & \multicolumn{1}{c}{\colorcellentropy{0.083}} & \multicolumn{1}{c}{\colorcellentropy{0.082}} & \colorcellentropy{0.080} \\ 
\hhline{-----|---}
\multirow{3}{*}{MLQA (vi)} & TOP & \multicolumn{1}{c}{\colorcellentropy{0.272}} & \multicolumn{1}{c}{\colorcellentropy{0.296}} & \colorcellentropy{0.267} & \multicolumn{1}{c}{\colorcellentropy{0.114}} & \multicolumn{1}{c}{\colorcellentropy{0.101}} & \colorcellentropy{0.107} \\ 
%\hhline{|~|-||---||---|}
 & MIDDLE & \multicolumn{1}{c}{\colorcellentropy{0.288}} & \multicolumn{1}{c}{\colorcellentropy{0.299}} & \colorcellentropy{0.295} & \multicolumn{1}{c}{\colorcellentropy{0.117}} & \multicolumn{1}{c}{\colorcellentropy{0.103}} & \colorcellentropy{0.120} \\ 
% \hhline{|~|-||---||---|}
 & BOTTOM & \multicolumn{1}{c}{\colorcellentropy{0.298}} & \multicolumn{1}{c}{\colorcellentropy{0.298}} & \colorcellentropy{0.313} & \multicolumn{1}{c}{\colorcellentropy{0.111}} & \multicolumn{1}{c}{\colorcellentropy{0.100}} & \colorcellentropy{0.112} \\ \hline
\end{tabular}%
}
\caption{Predictive entropy values for two models Llama3.1 8B and Qwen2.5 7B.}
\label{tab:entropy_full2}
\end{table*}

\clearpage

\section{Prompts}
\label{app:prompts}

Our user--prompts for LLM-inference in terms of context placement strategies can be seen in Tables \ref{tab:prompts_wos}, \ref{tab:prompts_ws}. For \textit{Aligned} and \textit{All Zero} strategies items in a contexts--list has a following format: "- [\{score\}] \{document\}". For \textit{No Scores} strategy items has the following format: "- \{document\}". As a system--prompt the same instruction was used for all languages (translated correspondingly): "You are an AI assistant who helps solve user issues.".

\begin{table*}[ht!]
%\vspace{1mm}
%\renewcommand{\arraystretch}{1.4}
\centering
\resizebox{\textwidth}{!}{%
\begin{tabular}{cl}
\hline
\textbf{Language} & \multicolumn{1}{c}{\textbf{User Prompt}} \\ \hline
\multirow{1}{*}{English} & 
\begin{minipage}{\textwidth}
\includegraphics[clip,trim={.08\textwidth} {.94\textheight} {.08\textwidth} 10mm, width=\textwidth]{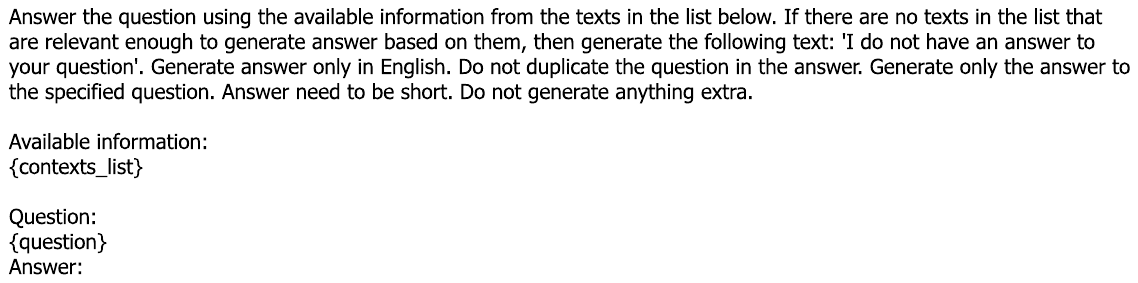}
\end{minipage}
\\ \hline

\multirow{1}{*}{Russian} & 
\begin{minipage}{\textwidth}
\includegraphics[clip,trim={.08\textwidth} {.94\textheight} {.08\textwidth} 10mm, width=\textwidth]{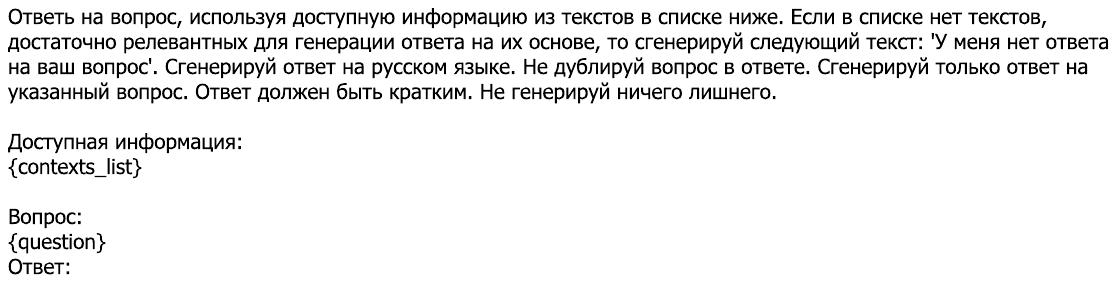}
\end{minipage}
\\ \hline

\multirow{1}{*}{German} & 
\begin{minipage}{\textwidth}
\includegraphics[clip,trim={.08\textwidth} {.93\textheight} {.08\textwidth} 10mm, width=\textwidth]{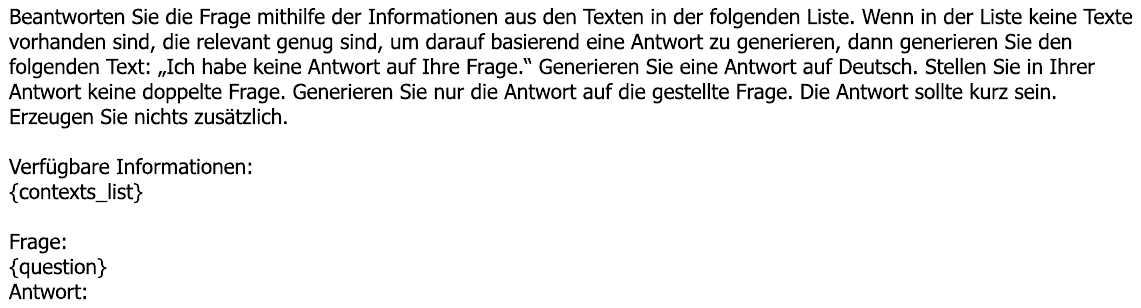}
\end{minipage}
\\ \hline

\multirow{1}{*}{Hindi} & 
\begin{minipage}{\textwidth}
\includegraphics[clip,trim={.08\textwidth} {.96\textheight} {.08\textwidth} 10mm, width=\textwidth]{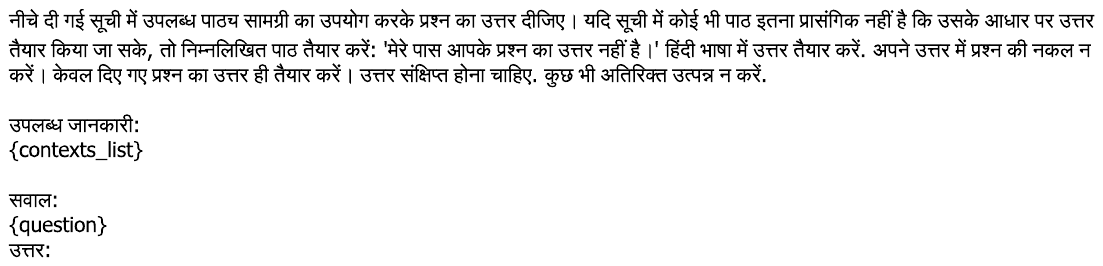}
\end{minipage}
\\ \hline

\multirow{1}{*}{Vietnamese} & 
\begin{minipage}{\textwidth}
\includegraphics[clip,trim={.08\textwidth} {.94\textheight} {.08\textwidth} 10mm, width=\textwidth]{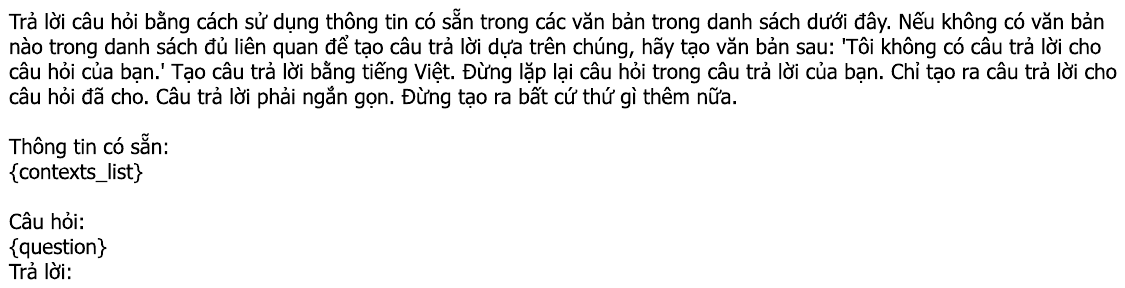}
\end{minipage}
\\ \hline
\end{tabular}%
}
\caption{User--prompts in five languages for LLM--inference in the \textit{No Scores} context placement strategy}
\label{tab:prompts_wos}
\end{table*}

\begin{table*}[ht!]
%\vspace{1mm}
%\renewcommand{\arraystretch}{1.25}
\centering
\resizebox{\textwidth}{!}{%
\begin{tabular}{cl}
\hline
\textbf{Language} & \multicolumn{1}{c}{\textbf{User Prompt}} \\ \hline
\multirow{1}{*}{English} & 
\begin{minipage}{\textwidth}
\includegraphics[clip,trim={.08\textwidth} {.89\textheight} {.08\textwidth} 10mm, width=\textwidth]{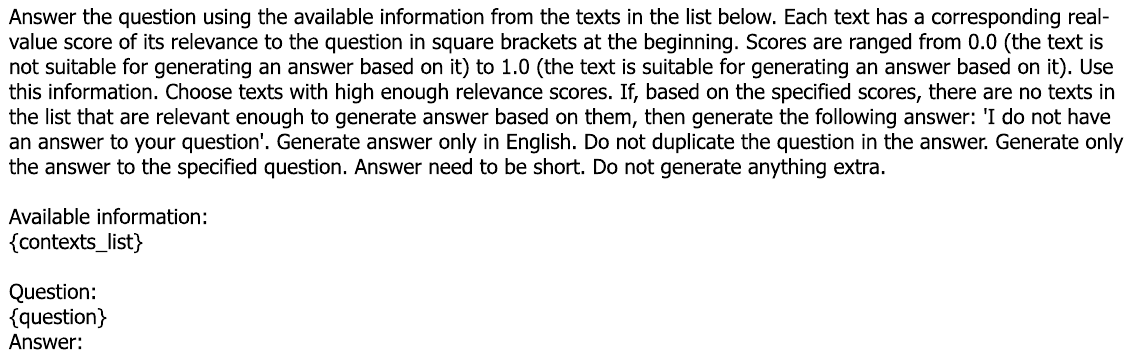}
\end{minipage}
\\ \hline

\multirow{1}{*}{Russian} & 
\begin{minipage}{\textwidth}
\includegraphics[clip,trim={.08\textwidth} {.87\textheight} {.08\textwidth} 10mm, width=\textwidth]{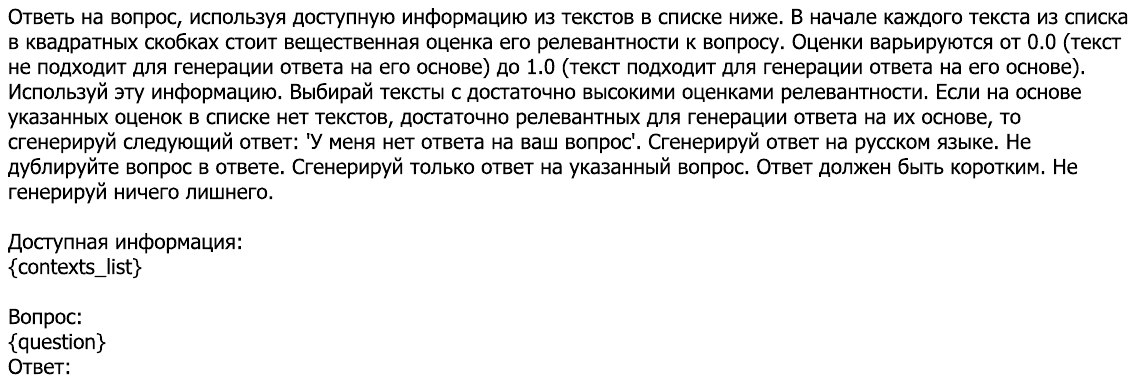}
\end{minipage}
\\ \hline

\multirow{1}{*}{German} & 
\begin{minipage}{\textwidth}
\includegraphics[clip,trim={.08\textwidth} {.85\textheight} {.08\textwidth} 10mm, width=\textwidth]{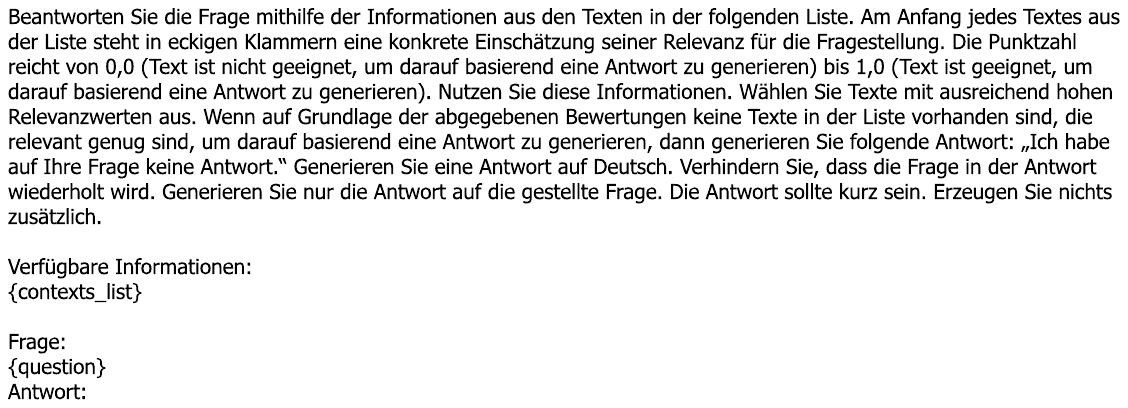}
\end{minipage}
\\ \hline

\multirow{1}{*}{Hindi} & 
\begin{minipage}{\textwidth}
\includegraphics[clip,trim={.08\textwidth} {.92\textheight} {.08\textwidth} 10mm, width=\textwidth,]{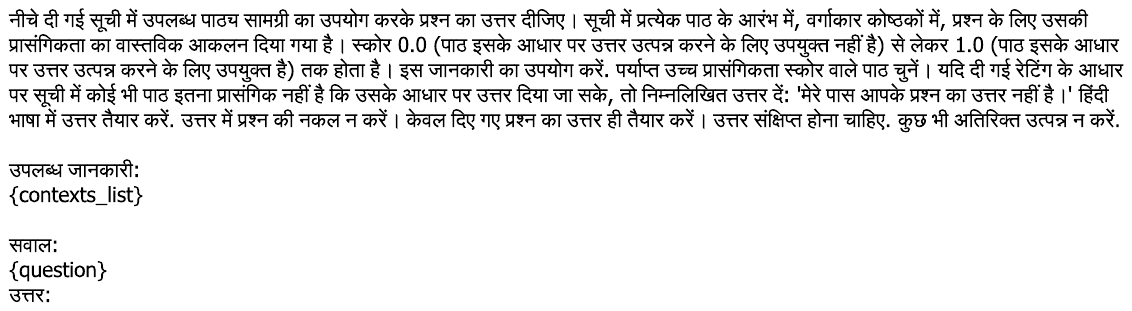}
\end{minipage}
\\ \hline

\multirow{1}{*}{Vietnamese} & 
\begin{minipage}{\textwidth}
\includegraphics[clip,trim={.08\textwidth} {.89\textheight} {.08\textwidth} 10mm, width=\textwidth]{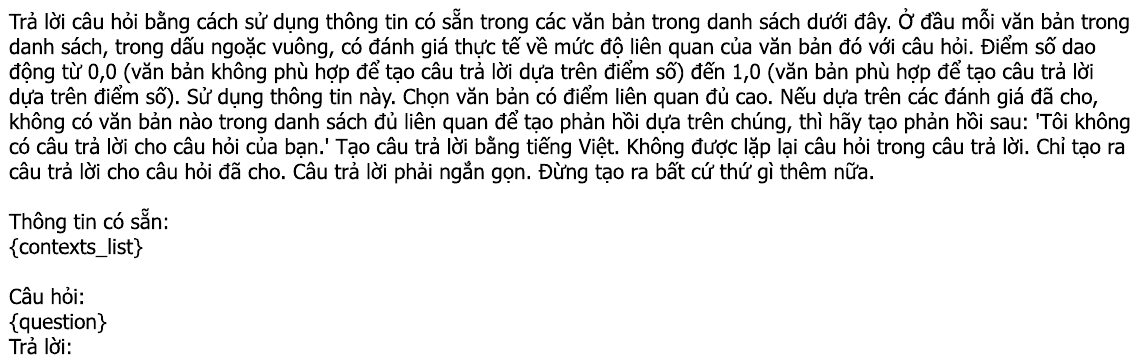}
\end{minipage}
\\ \hline
\end{tabular}%
}
\caption{User--prompts in five languages for LLM--inference in the \textit{Aligned} and \textit{All Zero} context placement strategies}
\label{tab:prompts_ws}
\end{table*}

\end{appendix}

\end{document}